\documentclass[letterpaper]{article} 

\usepackage{aaai2026}  
\usepackage{times}  
\usepackage{helvet}  
\usepackage{courier}  
\usepackage[hyphens]{url}  
\usepackage{graphicx} 
\usepackage{natbib}  
\usepackage{caption} 
\usepackage{algorithm}

\usepackage{newfloat}
\usepackage{listings}
\DeclareCaptionStyle{ruled}{labelfont=normalfont,labelsep=colon,strut=off} 
\floatstyle{ruled}
\newfloat{listing}{tb}{lst}{}
\floatname{listing}{Listing}
\usepackage{amsmath}
\usepackage{algpseudocode}
\usepackage{xcolor}
\algrenewcommand{\algorithmiccomment}[1]{ \textcolor{gray}{\# #1 \hfill } }
\usepackage{amsfonts}
\usepackage{adjustbox}

\title{Self-Improvement for Fast, High-Quality Plan Generation}
\author {
    Robert Gieselmann,
    Henrike von Huelsen,
    Mihai Samson,
    Marie-Christine Meyer,\\
    Dariusz Piotrowski, 
    Oleksandr Radomskyi, 
    Justin Okamoto, 
    Turan Gojayev, 
    Michael Painter, 
    Gavin Brown,
    Federico Pecora, 
    Jeremy L. Wyatt
}
\affiliations {
    Amazon\\
    \{robgie, vohenrik, samsonmf, mariemea, piotrod, radomole, justmoto, \\ otugojay, painterm, brorgav, fpecora, jeremywy\}@amazon.com
}

\nocopyright

\begin{document}

\maketitle

\begin{abstract}
Generative models trained on synthetic plan data are a promising approach to generalized planning. Recent work has focused on finding any valid plan, rather than a high-quality solution. We address the challenge of producing high-quality plans, a computationally hard problem, in sub-exponential time. First, we demonstrate that, given optimal data, a decoder-only transformer can generate high-quality plans for unseen problem instances. Second, we show how to self-improve an initial model trained on sub-optimal data. Each round of self-improvement combines multiple model calls with graph search to generate improved plans, used for model fine-tuning. An experimental study on four domains: Blocksworld, Logistics, Labyrinth, and Sokoban, shows on average a 30\% reduction in plan length over the source symbolic planner, with over 80\% of plans being optimal, where the optimum is known. Plan quality is further improved by inference-time search. The model's latency scales sub-exponentially in contrast to the satisficing and optimal symbolic planners to which we compare. Together, these results suggest that self-improvement with generative models offers a scalable approach for high-quality plan generation.
\end{abstract}


\section{Introduction}

Generative models have emerged as a useful tool for automated planning, providing a mechanism for retrieving and generalizing previously observed action sequences. Autoregressive models like transformers can quickly generate complete plan sequences that are then validated against formal verifiers like VAL \cite{howey2004val}. Recent work \cite{rossetti2024learning} has shown that this approach can achieve high completion rates across classical planning domains when trained on large datasets of valid plans. These generative models enable a form of \emph{generalized planning} \citep{hu2011generalized}: once trained, they solve new problems within the same domain without requiring per-instance search.

Prior work such as  \citet{rossetti2024learning} has focused primarily on plan validity rather than optimality. Generating optimal plans is computationally much harder than finding a valid solution. For instance, while Blocksworld admits linear-time algorithms, optimal planning in this domain is NP-hard \cite{SLANEY2001119, HELMERT2003219}. This computational barrier makes it infeasible to generate large optimal training datasets, fundamentally limiting generative models to the quality of their training data.

To overcome this quality ceiling, we present \emph{SiGPlan} (\textbf{S}elf-\textbf{i}mprovement for \textbf{G}eneralized \textbf{Plan}ning), a framework that iteratively refines a generative model to produce near-optimal plans without requiring optimal training data. Unlike search-based self-improvement frameworks such as AlphaGo \cite{silver2016mastering}, \emph{SiGPlan} is designed to leverage the rapid candidate plan synthesis that generative models excel at. Starting from a model pretrained on sub-optimal plans, our self-improvement loop alternates between (1) discovering improved plans by combining model predictions with search, and (2) model finetuning on these improved labels. \emph{SiGPlan} automatically derives a near-optimal, domain-specific generalized planner from sub-optimal training data.

We first establish an upper bound on generative model performance by training on optimal plans for Blocksworld, achieving a 78.9\% optimality rate for problems with up to 100 blocks. We then evaluate \emph{SiGPlan} on four domains: Blocksworld, Logistics, Labyrinth, and Sokoban. Across all domains, \emph{SiGPlan} substantially improves upon the initial model, achieving plan length reductions of 39.3\% on Blocksworld and 38.0\% on Labyrinth, demonstrating that self-improvement can break through the supervised learning quality ceiling. Moreover, \emph{SiGPlan} is computationally efficient, solving problems in seconds while optimal symbolic planners may require hours or days.

In summary, our main contributions are:
\begin{itemize}
\item An empirical analysis establishing the performance of generative models trained on optimal plans for Blocksworld with up to 100 blocks.
\item \emph{SiGPlan}, a self-improvement framework that iteratively refines generative planning models to approach optimal performance without requiring optimal training data.
\item A benchmark across four domains demonstrating that self-improvement enables models to substantially exceed their initial training data quality, producing generalized planners with near-optimal efficiency.
\item A runtime analysis showing that learned models achieve orders-of-magnitude speedups over optimal symbolic planners while maintaining near-optimal plan quality.
\end{itemize}

\section{Background and Related Work}

\paragraph{Automated Planning in PDDL.} Automated planning \cite{nau04automated} finds action sequences that transform an initial state into one satisfying goal conditions. The Planning Domain Definition Language (PDDL) \cite{mcdermott1998pddl,fox2003pddl21} provides a standardized formalism for planning problems. Classical PDDL domains consist of predicates, object types, and operators with preconditions and effects. Problem instances include objects, an initial state, and goal specifications. States are sets of atoms (propositional statements combining predicates with object tuples) that hold true in specific configurations. PDDL has become the standard representation in planning research and serves as the foundation for competitions like the International Planning Competition\footnote{https://ipc2023-classical.github.io/}. We use PDDL as our main formalism, providing a general recipe to tokenize planning problems for training and sequence generation with transformer models.

\paragraph{Sequence Modeling with Transformers.} Transformers \cite{vaswani2017attention} are a neural network architecture that have revolutionized sequence modeling across multiple domains. They leverage self-attention mechanisms to capture dependencies between all positions simultaneously, enabling efficient parallel computation. Multi-head attention allows joint attention to information from different representation subspaces, while position encodings preserve sequential information. The Generative Pretrained Transformer (GPT) \cite{radford2019language} is designed for sequence generation, predicting sequences autoregressively by feeding generated tokens back as input. It scales effectively through parallelizable training via causal masking and inference speedup via kv-caching. The GPT architecture underpins modern large language models and has been adopted in domains where applications can be formulated as sequence prediction, e.g., offline reinforcement learning \cite{janner2021offline} or Boolean satisfiability \cite{pan2025can}.

\paragraph{Pre-trained LLMs for Planning.} 
Recent work has explored leveraging pre-trained language transformers for automated planning. \citet{pallagani2023plansformer} rely on an LLM that was pretrained for coding tasks and finetuned it for optimal planning in several PDDL benchmark domains. They demonstrated that for small-scale problems (e.g., 2-5 blocks in Blocksworld), their approach outperformed the base LLM in both completion rate and plan quality. Moreover, the model generated plans significantly faster than an optimal symbolic solver. Other research has focused on LLMs for planning from natural language descriptions, including auto-formalizing problems into PDDL \citep{liu2023llmp}, generating world models \citep{guan23leveraging, katz24thought}, and synthesizing domain-specific planning policies as Python code \citep{silver2024generalized}. Despite these promising results, there remains skepticism about large pre-trained transformers due to the high inference costs and limited reliability \citep{kambhampati2024position}.

\paragraph{Training Planning Generative Models from Scratch.} Smaller, domain-specialized transformers have recently emerged as effective alternatives for automated plan generation. \emph{PlanGPT} \citep{rossetti2024learning} uses a large corpus of valid plans generated by a domain-independent planner to train domain-specific GPT transformers from scratch. The authors developed a specialized tokenization scheme representing actions as sequences of predicate and object tokens, achieving notable results across domains, such as 100\% plan completion on unseen Blocksworld test problems with up to 20 blocks. In follow-up work, \citet{rossetti2024enhancing} integrated an action validator into the sequence generation pipeline to prevent infeasible plans and thereby improved completion rates. \citet{tummolo2024integrating} introduce an additional processing step which repairs invalid plans by using them as the initialization for a local planning algorithm. \citet{fritzsche2025symmetry} proposed a symmetry-aware contrastive learning objective to enhance \emph{PlanGPT}'s ability to generalize to unseen configurations, especially problem instances with numbers of objects that were not seen during training. 

\emph{PlanGPT} is neither trained on optimal data nor evaluated in terms of the length of the generated plans. To address this gap, we devise a method that reduces plan lengths over time through an iterative process that alternates between search and learning.

\paragraph{Self-Improvement via Search.} 
The idea of combining search and learning in a loop of self-improvement dates back to Arthur Samuel's checkers program \citep{samuelscheckers}. \citet{JABBARIARFAEE20112075} applied this concept to classical planning with heuristic search. Their method alternates between graph search and training a shallow neural network heuristic, where the current model guides search while solved instances update the model for subsequent iterations.
Self-improvement with deep neural networks gained widespread attention with \emph{AlphaGo} \citep{silver2016mastering}, which integrated Monte Carlo Tree Search with deep policy and value networks.
\citet{groshev2018learning} used a neural network as a learned policy and heuristic for generalized planning. Under the name \emph{leapfrogging}, they introduce a system where model-guided A* search provides training labels for a curriculum of increasingly harder problem instances. In contrast, our work centers around autoregressive generative models, while focusing on improving plan optimality while maintaining fast runtimes.
Recently, numerous studies have adopted search-based self-improvement frameworks to enhance transformers \citep{lehnert2024beyond, zhang2024rest, alphamath}.
Our work applies this strategy to automated planning, leveraging the efficiency of smaller specialized models as demonstrated by \emph{PlanGPT} \cite{rossetti2024learning}.

\section{Training an Optimal Plan Generator}\label{sec:GM}
Our aim is to train a generative model that can generate (near-)optimal plans on unseen PDDL problem instances from the same distribution as the training data.
In this section, we show that a generative model trained on optimal 
plans can generate optimal solutions for the majority of unseen test instances,
and near-optimal solutions for the rest. 

\subsection{Problem Definition}\label{sec:problemdefinition}
We consider single-agent planning problems defined over a finite set of objects $\mathcal{O}$ and predicates $\mathcal{P}$. Each predicate $p \in \mathcal{P}$ has arity $\mathrm{arity}$(p). The set of grounded atoms is $\mathcal{X} = \bigcup_{p \in \mathcal{P}} \left\{\, p(o_1, \ldots, o_{\mathrm{arity}(p)}) \mid o_i \in \mathcal{O} \,\right\}.$  The state space is $\mathcal{S} = 2^\mathcal{X}$, where each state $s \in \mathcal{S}$ is the set of propositions that are true in that world configuration. The action space $\mathcal{A}$ consists of all grounded operators $a=\langle \mathrm{pre}(a), \mathrm{add}(a), \mathrm{del}(a) \rangle$, where $\mathrm{pre}(a)$, $\mathrm{add}(a)$ and $\mathrm{del}(a)$ are subsets of $\mathcal{X}$ defining preconditions, add, and delete effects,  respectively. The deterministic transition function is $F(s,a) = (s \setminus  \mathrm{del}(a)) \cup \mathrm{add}(a)$ which applies when $\mathrm{pre}(a) \subseteq s$. The goal space $\mathcal{G} \subseteq \mathcal{S}$ consists of partial states $g \in \mathcal{G}$, where a state $s$ satisfies a goal $g$ if $g \subseteq s$. A valid plan is a sequence of actions $\tau = ( a_0, \ldots, a_{T-1} )$ such that, starting from an initial state $s_0 \in \mathcal{S}$ and applying transitions $s_{t+1} = F(s_t, a_t)$, the resulting state $s_T$ satisfies $g \subseteq s_T $ for a given goal $g \in \mathcal{G}$. Among all valid plans, we define the optimal solution as the one that minimizes the plan length. 

We define $\mathcal{I} = \mathcal{S} \times \mathcal{G}$ as the space of problem instances, where each instance $(s_0,g) \in \mathcal{I}$ consists of an initial state $s_0 \in \mathcal{S}$ and a goal $g \in \mathcal{G}$. These problem instances are generated from an unknown distribution $P_\mathcal{I}$. Given a planning domain, our objective is to learn a fast plan generator that generalizes across problems from $P_\mathcal{I}$ while minimizing plan length. To achieve this, we assume access to a training set of problem instances $\mathcal{D}_\mathcal{I}$ generated by sampling from $P_\mathcal{I}$.

\subsection{Generative Model}\label{sec:GM:gm}
We train a generative model $\pi$ which maps from a starting state $s_0 \in \mathcal{S}$, goal $g \in \mathcal{G}$, and action history $a_{<t} \in \mathcal{A}^t$ to a distribution over next actions: $\pi(a_t | s_0, g, a_{<t})$. Following \citet{rossetti2024learning}, $\pi$ is implemented as a decoder-only transformer \cite{radford2019language} that operates at the token level. The model generates action sequences token-by-token through iterative forward passes, handling the combinatorially large space of grounded actions with a fixed-size vocabulary. For instance, in Blocksworld, the grounded action \texttt{unstack b1 b2} (unstacking block \texttt{b1} from block \texttt{b2}) is represented as a token sequence: the \texttt{unstack} operator token followed by tokens for \texttt{b1} and \texttt{b2}. The model operates autoregressively, appending each generated token to its input sequence after every prediction step.

\begin{figure}[t]
    \centering
    \includegraphics[width=\columnwidth]{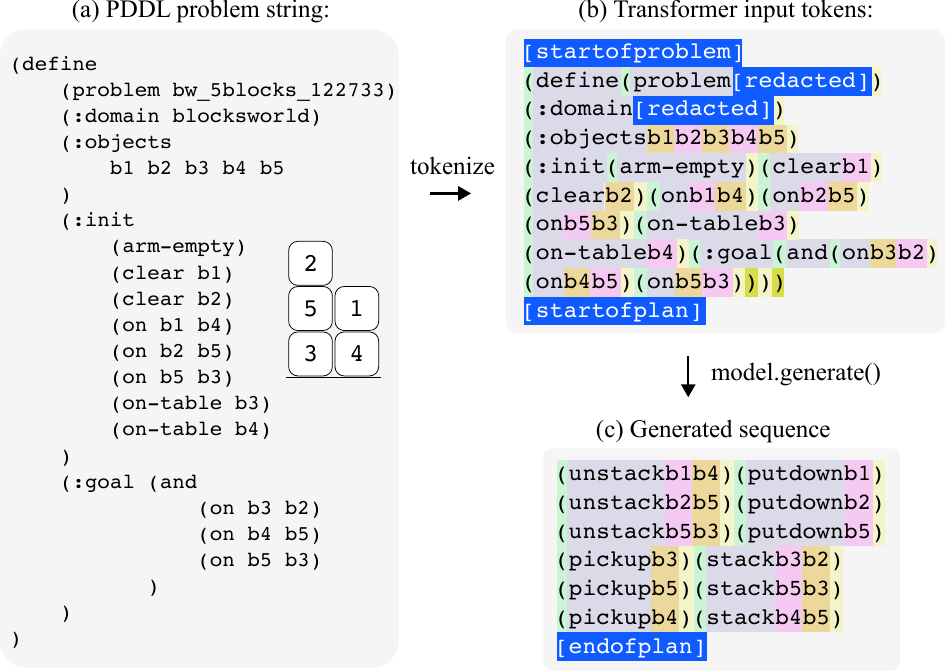}
    \caption{Illustration of our tokenization scheme.}
    \label{fig:tokenization}
\end{figure}

The vocabulary is constructed from PDDL domain and problem files, comprising predicate names, object identifiers, types, and special delimiter tokens. For each domain, we define a maximum number of objects per type. Figure~\ref{fig:tokenization} illustrates the tokenization scheme during inference on a Blocksworld problem instance. The PDDL problem file (\ref{fig:tokenization}a) is translated into a sequence of tokens (\ref{fig:tokenization}b). Special tokens such as \texttt{[startofproblem]}, \texttt{[startofplan]}, and \texttt{[endofplan]}  structure the input. The model then generates a plan as a sequence of action tokens, terminated by \texttt{[endofplan]} (\ref{fig:tokenization}c).  Generated plans are validated by splitting sequences into actions and verifying syntactic and semantic validity using VAL \citep{howey2004val}. 

During training, the model is optimized using the standard cross-entropy loss for next-token prediction. 

\subsection{Training from Optimal Plan Data}\label{sec:GM:optimality_experiments}

Obtaining optimal training data with domain-independent classical solvers is usually prohibitively expensive. However, for the Blocksworld domain, a domain-specific optimal solver was introduced by \citet{SLANEY2001119}. Blocksworld is a common benchmark domain where blocks must be rearranged from an initial configuration to a goal configuration by stacking and unstacking one block at a time.
We generate a training dataset of 999,270 Blocksworld problem instances of 3 to 100 blocks with optimal plans computed by the Slaney-Thiébaux solver. To satisfy the assumptions of the solver, problem instances are generated such that all objects appear in the goal specification.
To establish whether efficient generation of (near-) optimal plans scales to larger problems, we evaluate performance of the trained model (Section \ref{sec:GM:gm}) on three held-out test sets: 1000 problems each with 3-10 blocks, 11-25 blocks, and 26-100 blocks, respectively. During inference, we generate plans in batches of size 10 and report the best solution. 

We compare the generative model against \emph{Fast Downward} (FD) \cite{helmert06}, a state-of-the-art domain-independent solver, in an optimal setting with a 20-minute timeout. To quantify plan quality (among solved instances), we report the percentage of optimal plans and the regret, which we define as the percentage length increase over the optimal solution: $\frac{\text{cost} - \text{cost}_\text{optimal}}{\text{cost}_\text{optimal}} \cdot 100\%$ (defined as $0\%$ when $\text{cost}_\text{optimal}{=}0$).

\begin{table}[ht]
\centering
\resizebox{\linewidth}{!}{%
\begin{tabular}{l|ccc}
\hline
 & 3-10 blocks & 11-25 blocks & 26-100 blocks \\
\hline
\multicolumn{4}{l}{\textbf{Generative Model (GPT2 transformer)}} \\
\hline
Compl. (\%) & 99.80 & 100.00 & 97.50 \\
Optimal (\%) & 100.0 & 98.70 & 39.18 \\
Regret (\%) & 0.00 (± 0.00) & 0.06 (± 0.02) & 1.65 (± 0.06) \\
Runtime (s) & 0.60 (± 0.01) & 1.85 (± 0.02) & 9.53 (± 0.15) \\
\hline
\multicolumn{4}{l}{\textbf{FD-optimal (${<}$20min})} \\
\hline
Compl. (\%) & 100.00 & 44.30 & 0.00 \\
Runtime (s) & 0.48 (± 0.06) & 116.70 (± 10.76) & - \\
\hline
\end{tabular}}
\caption{Results of scalability analyses for 3 Blocksworld test sets containing 1000 unique problems each. For regret and runtime we report  mean ± std. error.}
\label{tab:scalability}
\end{table}

The results in Table \ref{tab:scalability} show that our model generates optimal or near-optimal plans across all three test sets (3-100 blocks).  
Even for the most challenging set with 26-100 block problem instances, the model maintained 97.5\% completion rate. While  the ratio of optimally solved instances dropped from almost 100\% for less than 25 blocks to 39\% for problems with 26-100 blocks, even the non-optimally solved problems were solved with less than 2\% regret on average.

Notably, for the test sets with more than 10 blocks, the model shows a significantly lower runtime than \emph{Fast-Downward}. While \emph{Fast Downward} could not solve any problem instance with over 25 blocks within 20 minutes, the model solved almost all problem instances with an average runtime of under 10 secs.
This underlines the prohibitively long runtime of classical (optimal) solvers, which grows exponentially with the problem complexity. 

The analysis shows that even for complex problems, the transformer model was able to generate a high percentage of optimal plans (78.9\% for 3-100 blocks), and near-optimal otherwise. The bottleneck, however, is computing optimal training data at scale, which is often infeasible in practice. This highlights the need for a method that can initially train on sub-optimal data and improve plan quality via self-refinement, which we address in the following section.

\section{A Self-improving Plan Generator}\label{sec:method}

\begin{figure*}[t]
    \centering
    \includegraphics[width=\textwidth]{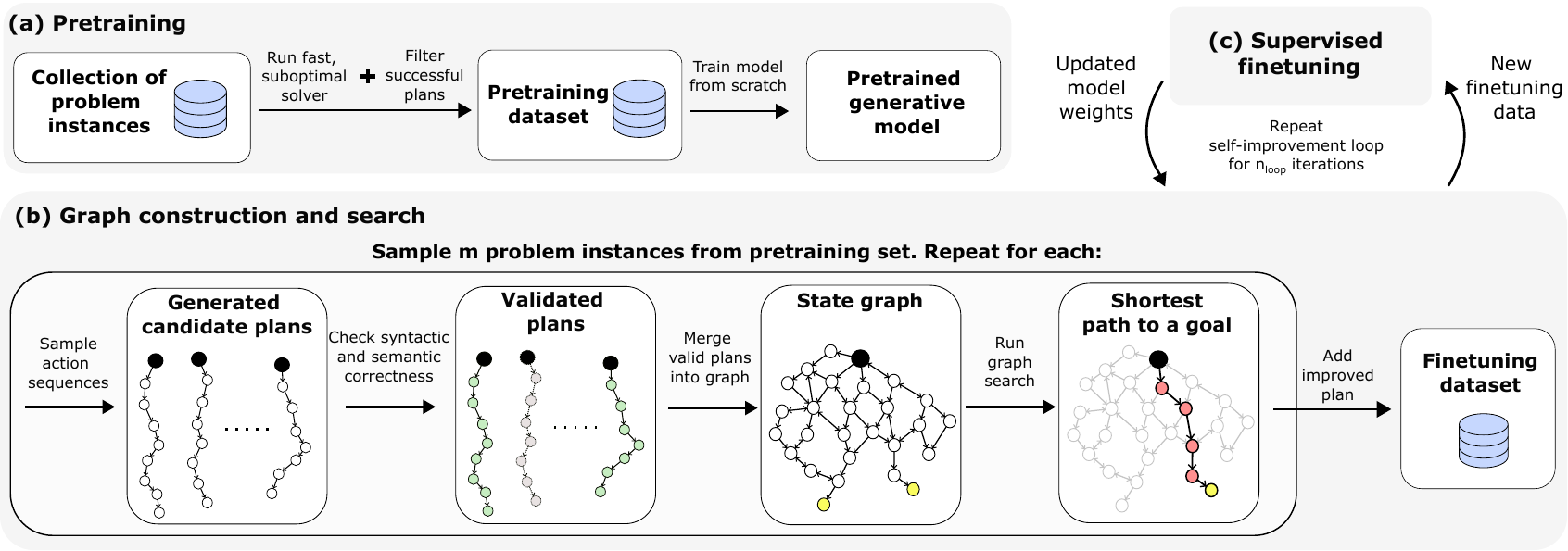}
    \caption{Overview of \emph{SiGPlan}. (a) The generative model is pretrained on plans generated by a domain-independent planner. (b) For a subset of $m$ problem instances, we sample candidate plans from our model, construct state graphs, and compute the shortest valid plans on these graphs. A finetuning dataset is created based on the best plan found for each problem in the subset. (c) The model is finetuned on the new data. Self-improvement iterates between (b) and (c) for a fixed number of steps.}
    \label{fig:overview}
\end{figure*}

We introduce \emph{SiGPlan} (\textbf{S}elf-\textbf{i}mprovement for \textbf{G}eneralized \textbf{Plan}ning), a self-improving planning framework that bootstraps from sub-optimal solvers. As shown in Figure \ref{fig:overview}, our approach has three components: pretraining, plan improvement via graph search, and model finetuning.

We train an initial generative model $\pi^0$ from scratch on a dataset of sub-optimal plans ($\mathcal{D}_\text{pretrain}$) generated by an off-the-shelf, domain-independent planner (Figure \ref{fig:overview}a). We then begin our self-improvement loop. For $m$ problem instances drawn randomly from $\mathcal{D}_\mathcal{I}$, we sample candidate plans from the current model to construct state graphs, then extract the shortest valid plans from these graphs via search (Figure \ref{fig:overview}b). The collection of improved plans forms a new dataset $\mathcal{D}_\text{finetune}^i$, which we use to finetune the generative model from the previous iteration. We choose finetuning over training from scratch as it requires significantly fewer new plans per iteration which is critical since generating these improved plans is computationally expensive due to numerous model calls. Preliminary experiments confirmed finetuning provides a better trade-off between plan quality improvement and computational cost. \emph{SiGPlan} runs for $n_\text{loop}$ steps, with each iteration $i$ using model $\pi^{i-1}$ to guide graph construction and extract improved training labels. 

Next, we detail our framework's building blocks: the generative model, graph search, and model finetuning.

\subsection{Graph Construction}

\begin{algorithm}[t]
\caption{Self-improvement Loop}
\label{alg:algorithm}
\textbf{Input}: Problem set $\mathcal{D}_\mathcal{I}$, initial data $\mathcal{D}_\text{pretrain}$, number of problem instances for finetuning $m$, number of candidate plans per problem $n$, self-improvement iterations 
$n_\text{loop}$\\
\textbf{Output}: $\pi$
\begin{algorithmic}[1]
\State \algorithmiccomment{Pretrain model on suboptimal data}
\State $\pi^0$ $\gets$ \texttt{pretrain}($\mathcal{D}_\text{pretrain}$)
\For{$i$ in $1 \ldots n_\text{loop}$}
    \State \algorithmiccomment{Sample subset of $m$ problem instances}
    \State $\mathcal{D}_\mathcal{I}^{i} \gets \texttt{sample\_problems}$($\mathcal{D}_\mathcal{I}$, num=$m$) 
    \State \algorithmiccomment{Generate set of candidate plans for each problem}
    \State $\{ \mathcal{T}_j \} \gets \texttt{sample\_plans}$($\mathcal{D}_\mathcal{I}^i$, $\pi^{i-1}$, num=$n$)
    \State \algorithmiccomment{Compute states (compile) and filter valid plans}
    \State $\{ \hat{\mathcal{T}}_j \} \gets \texttt{compile\_and\_filter}$($\mathcal{D}^i_\mathcal{I}$, $\{ \mathcal{T}_j \}$)
    \State \algorithmiccomment{Create set of state graphs}
    \State $\{ G_j \} \gets \texttt{construct\_graphs}$($\{ \hat{\mathcal{T}}_j \}$)
    \State \algorithmiccomment{Search graphs and extract shortest plan data}
    \State  $\mathcal{D}_\text{finetune}^i \gets \texttt{search\_and\_extract\_data}$($ \{G_j \} $)
    \State \algorithmiccomment{Finetune plan generator}
    \State $\pi^i$ $\gets$ \texttt{finetune}($\pi^{i-1}$, $\mathcal{D}^i_\text{finetune}$)
\EndFor\\
\Return $\pi$
\end{algorithmic}
\end{algorithm}

A critical step within \emph{SiGPlan} is to compute better plans for model finetuning. In summary, we first sample plans from our generative model and combine these to a graph representation of states and actions. To obtain improved training labels, the shortest path that satisfies the goal is extracted from each graph (Figure \ref{fig:overview}b). The procedure is formalized in Algorithm \ref{alg:algorithm}.

At each iteration $i{=}1,\dots, n_\text{loop}$, we start by drawing $m$ problem instances uniformly from $\mathcal{D}_\mathcal{I}$. The resulting problem subset is denoted by $\mathcal{D}^i_\mathcal{I}$. For each problem $(s_0,g) \in \mathcal{D}^i_\mathcal{I}$, we use our most recent generative model $\pi^{i-1}$ to synthesize $n$ candidate plans $\{\tau_k\}_{k=1}^n$, where each $\tau_k {=} \{ a_0, a_1, \ldots, a_{T_k-1} \}$ is a sequence of actions with individual length $T_k$. This sampling procedure can be performed efficiently by leveraging batch computation on GPUs. For each problem, the autoregressive generation stops either if the \texttt{[endofplan]} token is generated or a maximum token limit is exhausted. Generated sequences that do not follow the correct syntax of PDDL are discarded. The set of remaining plans for each problem instance  $j \in \{1,\ldots,m \}$ is denoted by $\mathcal{T}_j$.

We continue by \emph{compiling} the candidate plans in $\mathcal{T}_j$, using the transition model $s_{t+1} {=} F(s_t, a_t)$ to translate them into state-action sequences $\hat{\tau}_k {=} (s_0, a_0, s_1, a_1, \ldots, s_{T_k})$. During this process, we filter out invalid plans and retain only those satisfying the following conditions:
\begin{enumerate}
\item Actions $a_t$ fulfill the operator preconditions at every step in the plan, i.e.,  $\mathrm{pre}(a_t) \subseteq s_t$
\item The plan achieves the goal $g$, i.e., $g \subseteq s_{T_k}$, where $s_{T_k} = F(s_{T_k-1}, a_{T_k-1})$.
\end{enumerate}

We collect all valid plans across the $n$ candidates for each problem $j \in \{1,\ldots,m \}$ into a set $\hat{\mathcal{T}}_j$.

Finally, we construct a set of state graphs $\{ G_j \}_{j=1}^m$, one per problem instance. Each directed graph $G_j =(V_j,E_j)$ is constructed such that $V_j$ contains all unique states appearing in $\hat{\mathcal{T}}_j$, and $E_j$ contains all unique state-action-state transitions present in $\hat{\mathcal{T}}_j$.

\subsection{Plan Harvesting and Finetuning}\label{sec:method:search}
After obtaining the graphs $\{ G_j \}$, we extract new plan labels to improve our model. For each graph $G_j$, we apply breadth-first search (BFS) to compute the shortest sequence of actions from the initial state to a goal. A new finetuning dataset $\mathcal{D}^i_\text{finetune}$ is created based on the shortest paths extracted from $\{ G_j \}$ for all $m$ problems. If no plan can be extracted from $G_j$ for a problem instance $j$ (i.e., when our generative model cannot synthesize valid plans), $j$ is excluded from $\mathcal{D}^i_\text{finetune}$.

To prevent performance regression over iterations, we maintain a global best solution cache across all iterations. For each problem in $\mathcal{D}^i_\text{finetune}$, if a shorter or equally short plan was found in previous iterations or during pretraining, we use that plan instead. This ensures $\mathcal{D}^i_\text{finetune}$ contains only the shortest plans observed so far for each solved problem.

Finally, we update the previous iteration model $\pi^{i-1}$ on $\mathcal{D}^i_\text{finetune}$ to obtain a new model $\pi^i$ via supervised learning on the (problem, plan) pairs. To avoid overfitting on this finetuning data, we use a lower learning rate and fewer gradient updates compared to the pretraining phase.

\section{Experiments}\label{sec:experiments}

\begin{table*}[htb]
\centering \renewcommand{\arraystretch}{1.2}
\resizebox{\textwidth}{!}{%
\begin{tabular}{l|cc|cc|cc|cc}
\hline
& \multicolumn{2}{c|}{\bf Blocksworld} & \multicolumn{2}{c|}{\bf Logistics} & \multicolumn{2}{c|}{\bf Labyrinth} & \multicolumn{2}{c}{\bf Sokoban} \\
Method & Compl. (\%) & Plan length & Compl. (\%) & Plan length & Compl. (\%) & Plan length & Compl. (\%) & Plan length \\
\hline
\emph{SiGPlan} ($n_\text{loop}{=}15$, $N{=}10$)   & 99.40     & 41.86 (± 0.57)    & 99.30     & 146.12 (± 3.20)       & 98.70     & 15.50 (± 0.23)    & 94.40     & 118.79 (± 2.97) \\
\emph{SiGPlan} ($n_\text{loop}{=}15$, $N{=}10^3$) & 99.90     & 40.91 (± 0.55)    & 99.70     & 145.25 (± 3.16)       & 99.30     & 14.80 (± 0.20)    & 98.60     & 121.98  (± 2.96)  \\
\emph{SiGPlan}+BFS ($n_\text{loop}{=}15$, $N{=}10^3$)   & 99.90  & 40.86 (± 0.55) & 99.70 & 145.20 (± 3.16) & 99.30 & 14.61 (± 0.18) & 98.60 & 121.91 (± 2.95)\\
\hline
$\pi^0$ ($N{=}10$)    & 99.80     & 68.98 (± 1.25)    & 99.30     & 161.85 (± 3.59)       & 96.70     & 25.01 (± 0.54)    & 94.60     & 134.26 (± 3.41) \\
$\pi^0$ ($N{=}10^3$)   & 99.90     & 59.85 (± 1.01)    & 99.90     & 157.10 (± 3.47)       &  100.00   & 18.13 (± 0.25)    & 99.20     & 132.52 (± 3.25) \\
$\pi^0$+BFS ($N{=}10^3$)         & 99.90 & 57.19 (± 0.94) & 99.90 & 156.62 (± 3.46) & 100.00 & 17.52 (± 0.23) & 99.20 & 130.69 (± 3.17) \\ \hline
\emph{FD-LAMA-first (${<}$20min) }                        & 100.00         & 79.43 (± 1.50)    & 100.00         & 165.32 (± 3.42)       & 100.00         & 25.78 (± 0.50)    & 100.00 & 149.05 (± 3.62) \\
\emph{FD-LAMA-anytime (${<}$20min) }                        & 100.00         & 44.85 (± 0.72)    &      100.00    &    160.93 (± 3.57)  & 100.00         &  16.25 (± 0.32)    & 100.00 & 143.43 (± 3.84) \\
\emph{FD-optimal} (${<}$48h)          & 63.00     & 30.12 (± 0.48)   & 16.90     & 27.31 (± 1.48)       & 100.00     & 12.82 (± 0.13)    & 48.90     & 47.03 (± 1.08)  \\
\hline
\end{tabular}}
\caption{Comparison of performance across domains. Plan length reports the mean (± standard error) for those problems which the respective method solved. $N$ indicates the number of candidate plans generated per problem.}
\label{tab:results}
\end{table*}

\begin{table*}[htb]
\centering \renewcommand{\arraystretch}{1.2}
\resizebox{\textwidth}{!}{%
\begin{tabular}{l|cc|cc|cc|cc}
\hline
& \multicolumn{2}{c|}{\bf Blocksworld} & \multicolumn{2}{c|}{\bf Logistics} & \multicolumn{2}{c|}{\bf Labyrinth} & \multicolumn{2}{c}{\bf Sokoban} \\
Method & Optimal plans & Plan length & Optimal plans & Plan length & Optimal plans & Plan length & Optimal plans & Plan length \\
\hline
\emph{SiGPlan} ($n_\text{loop}{=}15$, $N{=}10$)        & \hfill 495~/~625 & 30.83 (± 0.50) & \hfill 95~/~169 & 28.29 (± 1.55)  & \hfill  619~/~955 & 15.24 (± 0.23) &  \hfill 406~/~468 & 47.14 (± 1.14) \\
\emph{SiGPlan} ($n_\text{loop}{=}15$, $N{=}10^3$)      & \hfill 600~/~625  & 30.24 (± 0.49) & \hfill 95~/~169 & 28.25 (± 1.55) & \hfill  657~/~955 & 14.51 (± 0.20)  & \hfill 439~/~468 & 46.76 (± 1.12) \\
\emph{SiGPlan}+BFS ($n_\text{loop}{=}15$, $N{=}10^3$)   & \hfill 606~/~625 & 30.20 (± 0.48) & \hfill 95~/~169 & 28.25 (± 1.55) & \hfill 662~/~955 & 14.37 (± 0.18) & \hfill 442~/~468 & 46.73 (± 1.12) \\
\hline
$\pi^0$ ($N{=}10$)         & \hfill 76~/~625 & 46.42 (± 1.00) & \hfill 62~/~169 & 30.31 (± 1.76) &  \hfill 284~/~955 & 24.80 (± 1.54)  & \hfill 195~/~468 & 52.24 (± 1.32) \\ 
$\pi^0$ ($N{=}10^3$)       & \hfill 103~/~625 & 40.83 (± 0.80) & \hfill 63~/~169 & 29.80 (± 1.71) &  \hfill 357~/~955 & 17.77 (± 0.25) & \hfill 247~/~468 & 50.12 (± 1.23) \\ 
$\pi^0$+BFS ($N{=}10^3$)         & \hfill 116~/~625 & 39.39 (± 0.76) & \hfill 63~/~169 & 29.76 (± 1.71) &  \hfill 389~/~955 & 17.22 (± 0.24) & \hfill 258~/~468 & 49.89 (± 1.22) \\ \hline
\emph{FD-LAMA-first (${<}$20min) }                              & \hfill  62~/~625 & 52.47 (± 1.22) & \hfill 47~/~169 & 31.21 (± 1.83) & \hfill 246~/~955 & 25.35 (± 0.51) & \hfill 118~/~468 & 57.25 (± 1.45) \\
\emph{FD-LAMA-anytime (${<}$20min) } & \hfill 595~/~625 & 30.50 (± 0.51) & \hfill 106~/~169 & 28.51 (± 1.62)  & \hfill 635~/~955 & 15.95 (± 0.32) & \hfill 421~/~468 & 47.13 (± 1.16)   \\ 
\emph{FD-optimal} (${<}$48h)                  & \hfill 625~/~625 & 30.14 (± 0.48)  & \hfill 169~/~169  & 27.31 (± 1.48) & \hfill 955~/~955 & 12.82 (± 0.13) & \hfill 468~/~468 & 46.46 (± 1.11) \\
\hline
\end{tabular}}
\caption{Comparison of performance \textbf{on the intersection of problems for which all methods found a solution}. `Optimal plans' gives the ratio between the number of problems for which the respective method found optimal plans, and the number of problems in the intersective set (e.g., out of 625 Blocksworld problems solved by all methods in the table).}
\label{tab:results_optimal}
\end{table*}

We benchmark \emph{SiGPlan} to demonstrate how it effectively enhances plan quality when starting from a model initially trained on suboptimal data. 
We evaluate on four domains:
\begin{itemize}
    \item \textbf{Blocksworld}: Introduced in Section \ref{sec:GM:optimality_experiments}. Here we use a more general setting where not all blocks are required to appear in the goal specification. Optimal planning in Blocksworld is NP-hard \cite{SLANEY2001119}.
    \item \textbf{Logistics}: A transportation domain where packages are delivered between airports and cities using trucks and airplanes. We consider instances with 1-50 cities, city sizes 1-5, 1-50 packages, 1-10 airplanes, and 1 truck per~city.
    \item \textbf{Labyrinth}: A navigation domain where each grid cell represents an intersection with different connectivity patterns (e.g., L-shaped or I-shaped). The agent can move between connected cells, as well as  shift entire rows or columns left/right/up/down, where cells pushed beyond one edge are reinserted at the opposite edge. We consider grid sizes of 3$\times$3 and 4$\times$4 cells.
    \item \textbf{Sokoban}: A puzzle domain where an agent pushes boxes to designated target locations while navigating around walls. We consider grid sizes of 5$\times$5 to 14$\times$14 cells with 1-10 boxes and 0-10 walls. Solving Sokoban is PSPACE-complete \cite{culberson98sokoban}.
\end{itemize}

\paragraph{Initial Datasets and Pretraining.} For each domain, we generate $|\mathcal{D}_\mathcal{I}|{=}|\mathcal{D}_\text{pretrain}|{=}10^5$ problem instances and compute corresponding suboptimal plans using \emph{FD-LAMA-first}~\cite{richter2010lama}. We pretrain a generative model $\pi^0$ on $\mathcal{D}_\text{pretrain}$ for 60 epochs on Blocksworld, 30 on Logistics, and 100 on Labyrinth and Sokoban. The checkpoint with the lowest validation loss (computed on 128 held-out problems) is selected as the starting point for \emph{SiGPlan}. We use 32 NVIDIA A100 GPUs (40\,GB) for training and 100 parallel GPU workers for plan sampling. Blocksworld and Labyrinth completed 15 iterations within one day, Logistics within two days, and Sokoban within 3.5 days.

\paragraph{Hyperparameters.} For all domains, we run $n_\text{loop}{=}15$ self-improvement iterations. At each iteration $i$, we sample $m{=}2000$ unique problem instances and construct graphs using $n{=}200$ generated plans per instance. For Logistics and Labyrinth, we use a transformer softmax temperature of 1, while for Blocksworld and Sokoban, a temperature of 2 provided a better trade-off between plan diversity and validity. We finetune on $\mathcal{D}_\text{finetune}^i$ for 30 epochs, empirically determined to improve the model without overfitting. The final checkpoint provides the model for the next iteration.

\paragraph{Evaluation Set and Methods.} We evaluate on 1000 held-out problems per domain. These instances are sampled from the set of problems solved by \emph{FD-LAMA-first} in less than 20 minutes. The following baselines are considered:
\begin{itemize}
\item \emph{SiGPlan}: Our finetuned generative model after $n_\text{loop}$ iterations of self-improvement 
\item \emph{$\pi^0$}: The generative model pretrained on $\mathcal{D}_\text{pretrain}$. 
\item \emph{FD-LAMA-first}: \emph{Fast Downward} with LAMA-first configuration. This symbolic planner generated $\mathcal{D}_\text{pretrain}$.
\item \emph{FD-LAMA-anytime}: \emph{Fast Downward} with LAMA anytime planning configuration (20min time limit).
\item \emph{FD-optimal}: \emph{Fast Downward} with LM-Cut heuristic for optimal planning (48h time limit).
\end{itemize}

\paragraph{Test-time compute.} For \emph{SiGPlan} and $\pi^0$, we evaluate two inference settings: $N{=}10$ and $N{=}10^3$, where $N$ is the number of candidate plans generated per problem instance. We report results for the shortest valid plan among the $N$ candidates. Additionally, we evaluate \emph{SiGPlan}+BFS and $\pi^0$+BFS, which augment the model predictions with graph construction and BFS (see Section~\ref{sec:method:search}).

\paragraph{Benchmark Results.} Table \ref{tab:results} shows \emph{SiGPlan}'s performance for all domains. Our method achieves high completion rates (98.6-99.9\%) across all tested domains, compared to \emph{FD-optimal} which solves 16.9-100.0\% of problems within its 48-hour timeout. Despite allowing \emph{FD-LAMA-anytime} a maximum compute time of 20 minutes, \emph{SiGPlan} still achieves lower average plan cost across all domains. \emph{SiGPlan} generates shorter plans compared to the pretrained generative model in all domains. Notably,  plan lengths are reduced by 39.3\% in the Blocksworld  domain (from 68.98 to 41.86 with $N{=}10$), and by 38.0\% for Labyrinth (from 25.01 to 15.50 with $N{=}10$). \emph{SiGPlan} ($N{=}10^3$) improves both completion rates and plan quality across all domains, showing benefits from increased sampling. \emph{SiGPlan}+BFS ($N{=}10^3$) yields only marginal improvements in plan length compared to \emph{SiGPlan} ($N{=}10^3$), i.e., the additional search components provide minimal benefits beyond what the finetuned model already achieves. The benefits of search are, however, larger for the pretrained model, as shown by the plan length improvements of $\pi^0$+BFS ($N{=}10^3$) over $\pi^0$ ($N{=}10^3$) — e.g., a reduction from 132.52 to 130.69 in Sokoban. These findings support that \emph{SiGPlan} scales to complex problems where optimal planners are intractable, while maintaining both high completion and short plans.
 
\begin{figure}[t]
    \centering
    \includegraphics[width=\linewidth]{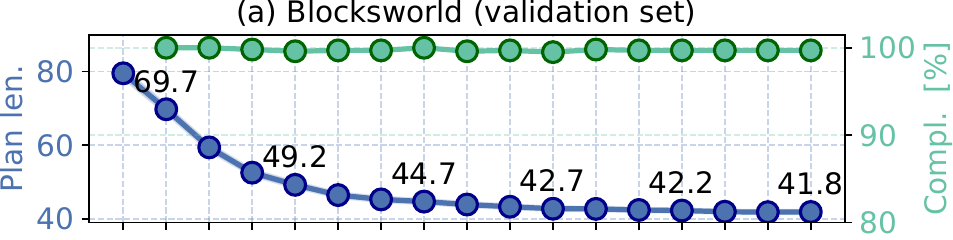}
    \includegraphics[width=\linewidth]{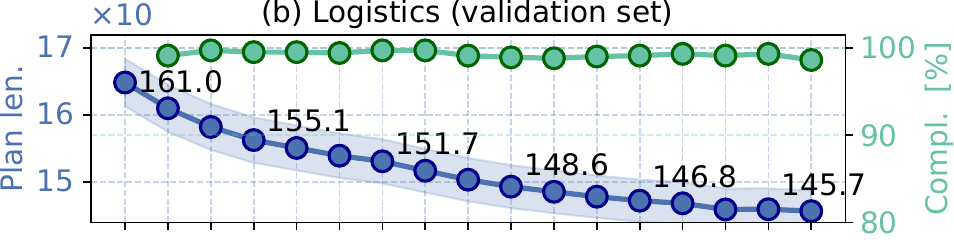}
    \includegraphics[width=\linewidth]{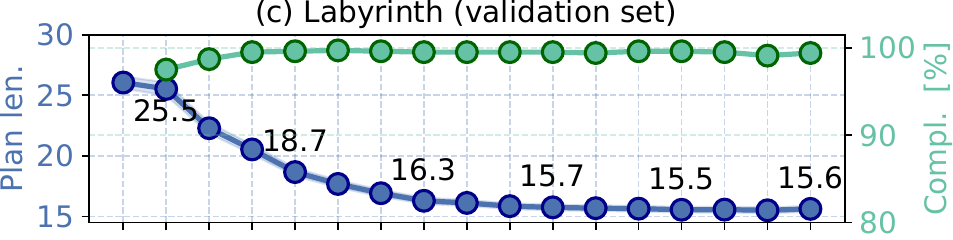}
    \includegraphics[width=\linewidth]{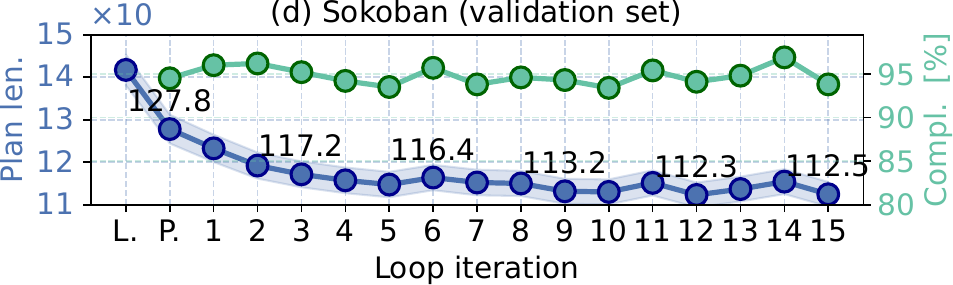}
    \caption{Plan length (mean ± std. error) and completion rates over self-improvement iterations (on 1000 unseen validation instances). `L.' refers to FD-LAMA-first and 'P.' to the pretrained model.}
    \label{fig:iterative_loop_validation}
\end{figure}

\paragraph{Convergence Behavior.} Figure~\ref{fig:iterative_loop_validation} shows average plan length and completion rates obtained with \emph{SiGPlan} ($N$=10) over self-improvement iterations on 1000 validation instances. Across all domains, \emph{SiGPlan} demonstrates rapid improvement in plan quality during the early iterations, with the most substantial gains occurring within the first 5 loop iterations. For Blocksworld and Labyrinth, the mean plan length decreases sharply from the pretrained model (69.7 to 41.8 and 25.5 to 15.6, respectively), with convergence stabilizing around iterations 8-10. Logistics exhibits more gradual but consistent model improvement throughout all 15 iterations (161.0 to 145.7). Sokoban shows initial improvement from the pretrained model (127.8) to iteration 6 (116.4), ultimately reaching 112.5 by iteration 15. Notably, completion rates remain stable at near 100\% for Blocksworld, Logistics, and Labyrinth throughout, indicating that the iterative refinement process improves plan quality without degradation.

\begin{figure}[t]
    \centering
    \includegraphics[width=\linewidth]{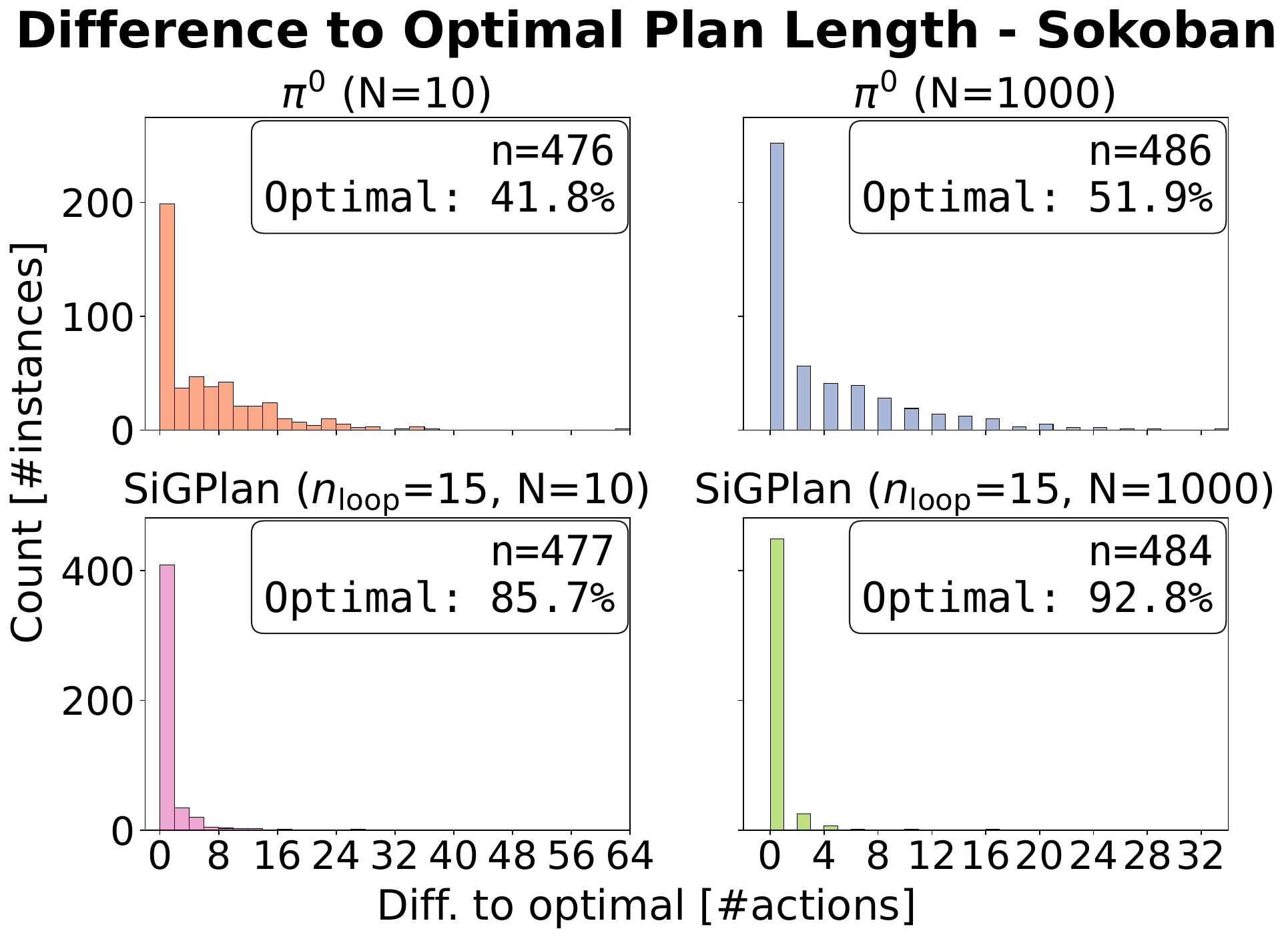}
    \includegraphics[width=\linewidth]{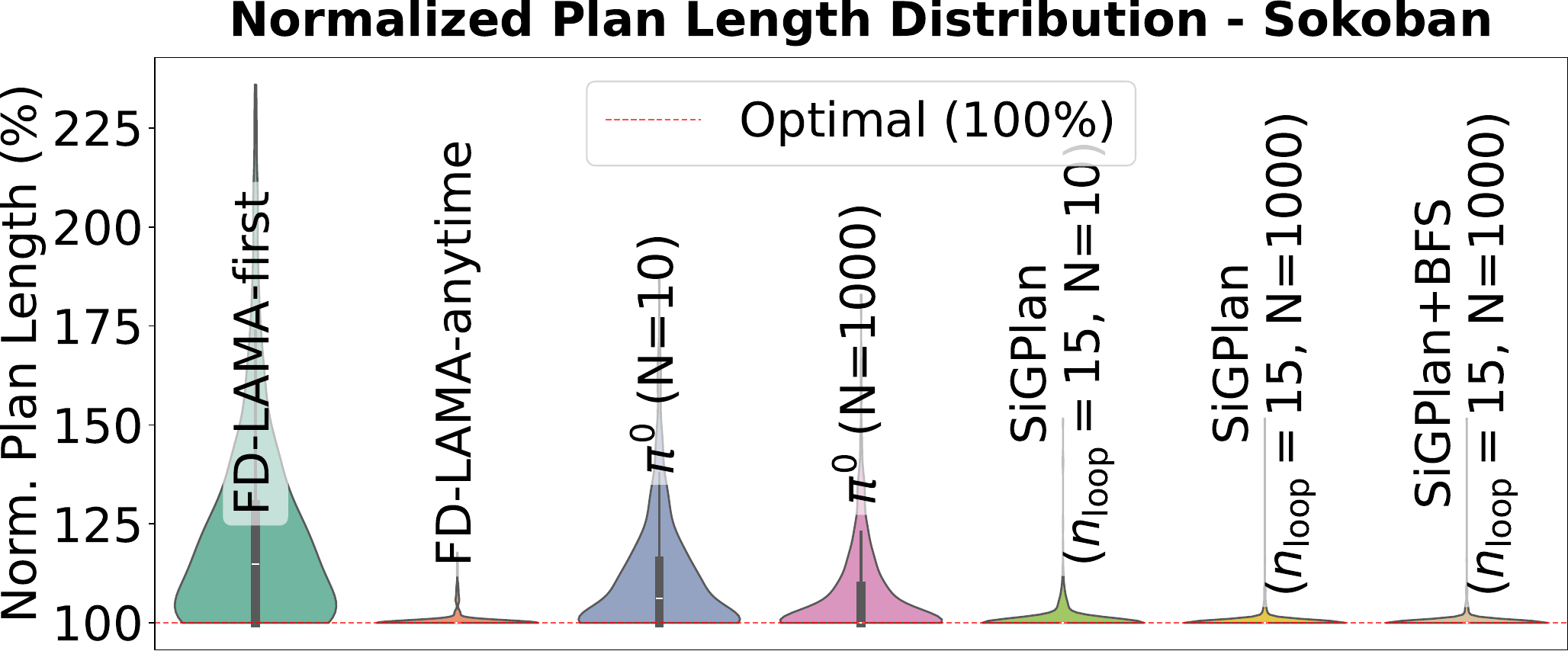}
    \caption{Plan length metrics for Sokoban. Histograms show the distribution of plan length differences to optimal. Violin plots show the distribution of normalized plan length, i.e., the plan length as a percentage of the optimal plan.}\label{fig:diff_to_opt_SOK}
\end{figure}

\paragraph{How close are we to optimal?} 
To obtain optimal baseline data, we ran \emph{FD-optimal} for 48 hours on all test sets. Table~\ref{tab:results_optimal} evaluates plan optimality on the subset of test problems solved by all methods. Using \emph{SiGPlan} ($n_\text{loop}{=}15$, $N{=}10^3$), we optimally solve 600/625 (96\%) of the Blocksworld problems, 95/169 (56.2\%) of Logistics problems, 657/955 (68.8\%) of Labyrinth problems, and 439/468 (93.8\%) of the Sokoban problems. When adding BFS at inference time, the number of optimal plans increases, e.g. to 606/625 (97\%) for Blocksworld and 442/468 (94.4\%) for Sokoban.
This substantially outperforms both the pretrained model $\pi^0$ and the classical  \emph{FD-LAMA-first} planner.
Notably, the mean plan lengths produced by \emph{SiGPlan} ($N{=}10^3$) closely approach those of \emph{FD-optimal}: 30.24 vs.~30.14 for Blocksworld, 28.25 vs.~27.31 for Logistics, and 46.76 vs.~46.46 for Sokoban. Figure \ref{fig:diff_to_opt_SOK} shows that most plans generated by \emph{SiGPlan} ($N{=}10^3$) in Sokoban are either optimal or few actions longer than the corresponding optimal plans. These results demonstrate that \emph{SiGPlan} drives the model toward near-optimal and often optimal solutions.

\begin{table}[ht]
\centering
\centering \renewcommand{\arraystretch}{1.2}
\resizebox{\columnwidth}{!}{%
\begin{tabular}{l|cccc}
\hline
\textbf{Method} & \textbf{Blocksw.} & \textbf{Logistics} & \textbf{Labyrinth} & \textbf{Sokoban}\\
\hline
\emph{SiGPlan} ($n_\text{loop}{=}15$, $N{=}10$) & \hfill 0.49 & \hfill 2.15 & \hfill 0.54 & \hfill 4.08\\
\emph{SiGPlan} ($n_\text{loop}{=}15$, $N{=}10^3$) & \hfill 21.21 & \hfill 112.91 & \hfill 27.81 & \hfill 407.66\\
\emph{FD-LAMA-first} (${<}$20min) & \hfill 0.39 & \hfill 7.32 & \hfill 270.15 & \hfill 12.70 \\
\emph{FD-LAMA-anyt.} (${<}$20min) & \hfill 985.82 & \hfill 765.71 & \hfill 935.28 & \hfill 796.18 \\
\emph{FD-optimal} (${<}$48h) & \hfill 13311.40 & \hfill 16670.74 & \hfill 3083.48 & \hfill 9762.96 \\
\hline
\end{tabular}}
\caption{Average solving times in seconds on test sets across different domains (computed on solved instances only).}
\label{tab:solving_times}
\end{table}

\begin{figure}[t]
    \centering
    \includegraphics[width=\linewidth]{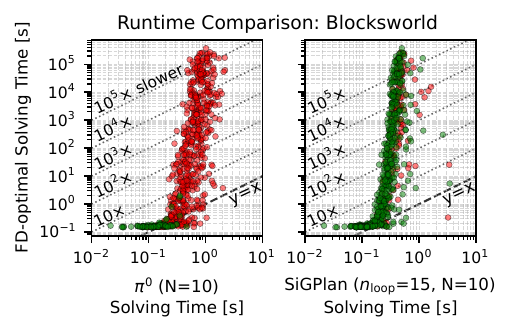}
    \caption{Runtime comparison between \emph{FD-optimal} and the generative models. Green dots represent instances where the model generated a plan with same length as \emph{FD-optimal}.}
    \label{fig:runtime_blockworld}
\end{figure}

\paragraph{Runtime Comparison.} Table~\ref{tab:solving_times} presents the average test runtimes across domains. As shown, \emph{SiGPlan} ($N{=}10$) requires on average only a few seconds of runtime. Despite requiring only a fraction of the time budget afforded to classical planners, \emph{SiGPlan} produces shorter plans across all domains (Table~\ref{tab:results}), making it a compelling approach for fast and high-quality plan generation. Figure~\ref{fig:runtime_blockworld} compares the solving times of $\pi^0$ ($N{=}10$) and \emph{SiGPlan} ($n_\text{loop}{=}15$, $N{=}10$) against \emph{FD-optimal} on Blocksworld. For this comparison, we ran \emph{FD-optimal} for 5 days on the test set, solving 643 instances. Both generative models solve problems in 0.1-10 seconds with consistent runtimes, while \emph{FD-optimal} requires anywhere from few seconds to over $10^5$ seconds ($>$27 hours). \emph{SiGPlan} demonstrates similar runtime characteristics to $\pi^0$, indicating that the improvements in plan quality come without additional computational cost at inference time. This highlights a key advantage of learned models: they provide near-optimal solutions at a fraction of the computational cost required by optimal symbolic planners.

\paragraph{Statistical Analysis.}
While Table 2 contains the descriptive statistics comparing our methods on the two performance metrics plan length and completion rate, inferential statistics is necessary to establish which of the observed differences are meaningful rather than due to random variation. 
We therefore conducted a statistical analysis (on commonly solved instances) comparing six specific conditions: (1) $\pi^0$ vs.~\emph{SiGPlan} with $N{=}10$, (2) $\pi^0$ vs.~\emph{SiGPlan} with $N{=}10^3$, (3) $\pi^0$ with $N{=}10$ vs.~$N{=}10^3$, (4) \emph{SiGPlan} with $N{=}10$ vs.~$N{=}10^3$, (5) \emph{SiGPlan} ($N{=}10^3$) vs.~\emph{SiGPlan+BFS} ($N{=}10^3$), and (6) \emph{SiGPlan+BFS} ($N{=}10^3$) vs.~\emph{FD-LAMA-anytime}.

For \textbf{Blocksworld}, all six comparisons of plan length were highly significant after Bonferroni-corrections ($p {<} 0.001$). \emph{SiGPlan} reduced mean plan length from 69.1 to 41.8 at $N{=}10$, and from 60.1 to 40.9 at $N{=}10^3$. Adding BFS improved plan quality significantly, but with a small effect size (\emph{SiGPlan}: 40.90 vs.~\emph{SiGPlan+BFS}: 40.85).  \emph{SiGPlan+BFS} substantially outperformed \emph{FD-LAMA-anytime} (40.9 vs.~44.9). Completion rates were consistently high (99.4--100\%) with no significant differences.

For \textbf{Logistics}, plan length differences were again highly significant ($p {<} 0.001$). \emph{SiGPlan} yielded mean reductions of 15.9 actions with $N{=}10$ (161.4 to 145.5) and 11.6 actions with $N{=}10^3$ (155.9 to 144.4). Increasing candidate plans to $N{=}10^3$ reduced plan length by 5.5 for $\pi^0$ and by 1.1 for \emph{SiGPlan}. Adding BFS improved plan quality significantly, but with a very small effect size (\emph{SiGPlan}: 144.37 vs.~\emph{SiGPlan+BFS}: 144.33), and \emph{SiGPlan+BFS} substantially outperformed \emph{FD-LAMA-anytime} (144.3 vs.~159.8). Completion rates showed no significant differences.

For \textbf{Labyrinth}, all differences in plan length are highly significant ($p {<} 0.001$) as well. \emph{SiGPlan} reduced mean plan length from 24.8 to 15.2 with $N{=}10$, and from 17.8 to 14.5 with $N{=}10^3$. Adding BFS improved plan quality significantly, but with a small effect size (\emph{SiGPlan}: 14.51 vs.~\emph{SiGPlan+BFS}: 14.37 actions), while \emph{SiGPlan+BFS} substantially outperformed \emph{FD-LAMA-anytime} (14.4 vs.~15.9 actions). In addition, completion rates varied meaningfully (96.7--100\%), with $N{=}10^3$ increasing completion from 96.7\% to 100\% ($p {<} 0.001$), and \emph{SiGPlan} improving completion from 96.7\% to 98.7\% ($p = 0.022$) with $N{=}10$. 

Plan length for \textbf{Sokoban} was again significantly different across all comparisons ($p {<} 0.001$). 
\emph{SiGPlan} reduced mean plan length by 14.6 with $N{=}10$ (130.5 vs.~115.9) and by 9.4 with $N{=}10^3$ (124.4 vs.~115.0). For $\pi^0$, increasing $N$ from $10$ to $1000$ reduced plan length by 6.2, and by 0.9 for \emph{SiGPlan}. Adding BFS improved plan quality significantly but only very  slightly (\emph{SiGPlan}: 115.04 vs.~\emph{SiGPlan+BFS}: 114.99). Lastly, \emph{SiGPlan+BFS} again outperformed \emph{FD-LAMA-anytime} (115.0 vs.~133.6). For completion rates, we found a significant effect of $N{=}10^3$ vs.~$N{=}10$ for both $\pi^0$ and \emph{SiGPlan} ($p {<} 0.001$). \emph{FD-LAMA-anytime} achieved significantly higher completion than \emph{SiGPlan+BFS} ($p {<} 0.001$), though all rates were high (94.4-100\%).

\section{Complexity Analysis}

\textbf{Plan validation} runs in $\mathcal{O}(T {\times} M)$ time, where $T$ is the maximum plan length and $M$ is the maximum number of precondition and effect atoms per action. \textbf{Graph construction} inserts each unique transition from valid plans using hash table lookups, running in time linear in the total number of transitions with memory complexity $\mathcal{O}(D {\times} N)$, where $D$ is the maximum rollout length and $N$ is the number of generated plans. \textbf{BFS} explores the graph from the initial state in $\mathcal{O}(V {+} E)$, where $V$ is the number of states and $E$ the number of transitions. For graphs constructed from at least one valid plan, every graph contains at least one goal state, and BFS's exhaustive exploration in order of increasing depth guarantees that the shortest path is found, making it both complete and optimal with respect to the constructed graph.
\section{Discussion \& Conclusions}\label{sec:discussion}

Finding efficient algorithms for optimal and near-optimal planning has been a long-standing goal in AI. We presented \emph{SiGPlan}, a self-improving generalized planning system that combines transformer models with symbolic graph search to iteratively improve plan quality, scaling favorably compared to traditional symbolic solvers.

The resulting system produces optimal plans for 72.8\% of the benchmark problems solved by \emph{FD-optimal}, with average per-domain latencies ranging from 0.49\,seconds (Blocksworld) to 4.08\,seconds (Sokoban). Inference-time scaling increases this proportion to 80.8\%, with latencies ranging from 21.21\,seconds (Blocksworld) to 407.66\,seconds (Sokoban), compared to 13311.40\,seconds and 9762.96\,seconds for \emph{FD-optimal}. Unlike \emph{FD-optimal}, model latency scales polynomially with the combined context and plan length. \emph{SiGPlan} represents an important step toward computationally efficient near-optimal planning.

\paragraph{Future Work.} Despite strong performance, several limitations remain. First, our approach requires an initial dataset of solutions, which may be hard to obtain for complex problems (e.g., Sokoban remains PSPACE-complete). Second, our exploration strategy is essentially on-policy, so convergence to optimal plans requires optimal actions to lie within the pretrained policy's support. Future work could incorporate off-policy exploration via A* or MCTS, and establish probabilistic guarantees for completeness or optimality.

\clearpage

\section*{Acknowledgments}
The authors would like to thank Marc Toussaint, Alexander Melkozerov, Aaron Parness, Sara Anwar, and Michael Zhang for their valuable support and contributions to this work.

\bibliography{aaai2026}

\clearpage

\appendix

\onecolumn

\section*{Appendix}

\section{Datasets}

For each of the four evaluated domains, we constructed datasets of PDDL problem instances and reference solutions. More information on the domain files can be found in Section \ref{app:domains}.

The dataset for each domain consists of 100k unique training instances, 1k unique validation instances (used to choose best iteration and/or checkpoint), and 1k unique test instances. All instances were solved by the Fast Downward planner in the LAMA-first setting with a timeout of 20 minutes. Only problem instances that were solved within this timeout appear in the dataset. 

\subsection{Blocksworld}
For the Blocksworld domain, we include problem instances from 3 to 25 blocks. All generated instances with up to 25 blocks were solved by Fast Downward in the LAMA-first setting with a timeout of 20 minutes. The distribution of instances in the dataset rises logarithmically with the number of blocks. A uniform distribution over block count is not possible in this domain, since there are only a few unique problem instances for lower block counts.
We used the problem instance generator from the PDDL generators repository \cite{seipp2022pddlgen}.

\subsection{Logistics}
We generated a dataset including problem instances with 1 -- 50 cities with a city size of 1 -- 5, 1 -- 50 packages and 1 -- 10 airplanes, and  1 truck per city. 
The problems were sampled using a uniform distribution over all combinations of object counts. Within 20 minutes, Fast Downward in the LAMA-first setting solved 99.97 \% of the instances. More instances were added to ensure a size of 100k/1k/1k.

With these parameters, our dataset spans a wider range of objects than most previous investigations in the domain. 
The Logistics domain was part of the IPC in 1998 and 2000 \cite{bacchus2001ipc}. The dataset from IPC 1998 shows similar numbers of objects to our dataset, while the IPC 2000 Logistics dataset contains a maximum of 15 packages and 5 cities. 
The Logistics domain is also included in the PlanBench benchmark dataset \cite{valmeekam2023planbench}, where the dataset includes a maximum of 24 objects distributed across the different types.

We re-implemented the problem instance generator in Python, closely following the provided C-implementation of \cite{seipp2022pddlgen}.

\subsection{Labyrinth}
The Labyrinth domain was part of the most recent IPC 2023 \cite{taitler2023ipc}. The competition dataset included 20 problem instances with map sizes between 5$\times$5 and 8$\times$8. We first generated a dataset with map sizes between 3$\times$3 and 12$\times$12 and up to 10 card rotations. On this dataset, Fast Downward in the LAMA-first setting solved only about 29\%. No problem instances with a map size bigger than 5$\times$5 were solved, and only about 15\% of the 5$\times$5 maps. Therefore, we only include maps of sizes 3$\times$3 and 4$\times$4 in the final dataset. 
To the best of our knowledge, the complexity of the Labyrinth domain has not been formally explored.
We use the problem instance generator provided within the IPC 2023 GitHub repository \cite{eifler2023labyrinth}.

\subsection{Sokoban}
The Sokoban domain was part of the IPC in 2008 and 2011. In both years, the datasets contained mostly instances with grid sizes between 7 and 15, and one map with grid size 30. The number of boxes reached up to 12.
Given grid sizes between 5$\times$5 and 30$\times$30, up to 12 boxes and up to 9 walls, Fast Downward in the LAMA-first setting solved only about 24\% of the problem instances. We therefore decided to lower the grid size to a maximum of 14$\times$14 with 1 to 10 boxes and up to 10 walls. We used the problem instance generator from the PDDL generators repository \cite{seipp2022pddlgen}.

\section{Training Details}
\label{app:training}

The model training for each domain was performed on 4 machines with 8 NVIDIA A100 GPUs with 40 GB VRAM. Pretraining took 2.5 hours for Blocksworld, 6.5 hours for Logistics, 4.5 hours for Labyrinth, and 48 hours for Sokoban. The hyperparameters for SiGPlan can be found in Tables \ref{tab:pre_hyperparameters} and \ref{tab:ft_hyperparameters} for the pretraining and finetuning, respectively. 

\begin{table}[h]
\centering
\renewcommand{\arraystretch}{1.2}
\begin{adjustbox}{max width=\columnwidth}
\begin{tabular}{l|c}
\hline
\textbf{Hyperparameter} & \textbf{Value (BW/Log/Lab/Sok)} \\
\hline
Number of transformer blocks ($n_\text{layer}$) & 12 \\
Number of attention heads ($n_\text{head}$) & 12 \\
Embedding dimension ($n_\text{embd}$) & 768  \\
Feedforward dimension ($n_\text{inner}$) & 3072 \\
Attention dropout rate & 0.1 \\
Embeddings dropout rate & 0.1 \\
\hline
Learning rate & 5e-5 \\
Learning rate scheduler & Cosine \\
Warmup steps & 1000 \\
Optimizer & AdamW \\
Batch size & 6/4/6/1 \\
\hline
Max sequence length & 14000 \\
Training epochs & 60/30/100/100 \\
\hline
\end{tabular}
\end{adjustbox}
\caption{Pretraining hyperparameters for the transformer decoder (GPT2).  Unless otherwise indicated, values are identical across domains. }
\label{tab:pre_hyperparameters}
\end{table}

\begin{table}[h]
\centering
\renewcommand{\arraystretch}{1.2}
\begin{adjustbox}{max width=\columnwidth}
\begin{tabular}{l|c}
\hline
\textbf{Hyperparameter} & \textbf{Value (BW/Log/Lab/Sok)} \\
\hline
Learning rate & 5e-6 \\
Learning rate scheduler & Cosine \\
Optimizer & AdamW \\
Batch size & 6/4/6/1 \\
\hline
Finetuning epochs & 30\\
Softmax temperature (inference) & 2/1/1/2 \\
Dataset size ($m$) & 2000 \\
Candidate plans per problem ($n$) & 200 \\
\hline
\end{tabular}
\end{adjustbox}
\caption{Finetuning hyperparameters for the transformer decoder (GPT2). Unless otherwise indicated, values are identical across domains.}
\label{tab:ft_hyperparameters}
\end{table}
\section{Comparing Solving Times}

Table \ref{tab:solving_times} presents the solving time statistics across the different domains for each method. The mean solving times were computed on the test sets (based on the subset of solved instances).

\begin{figure*}[htb]
    \centering
    \includegraphics[width=0.75\linewidth]{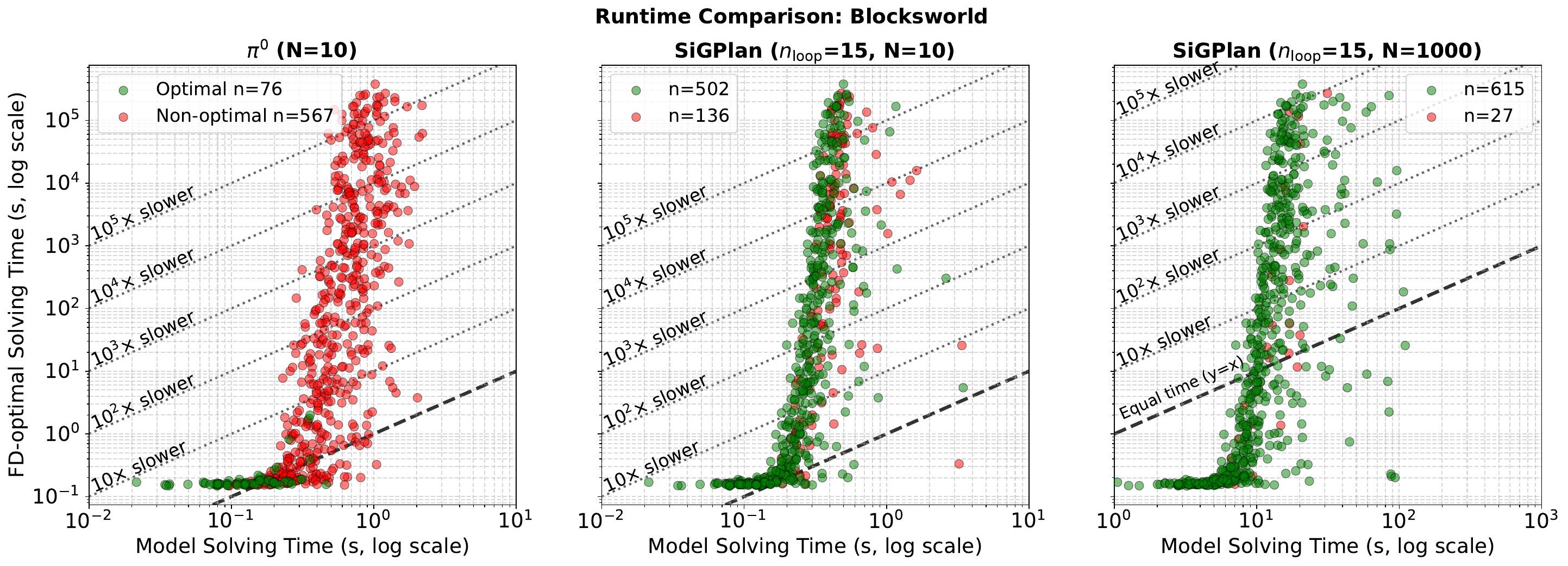}
    \includegraphics[width=0.75\linewidth]{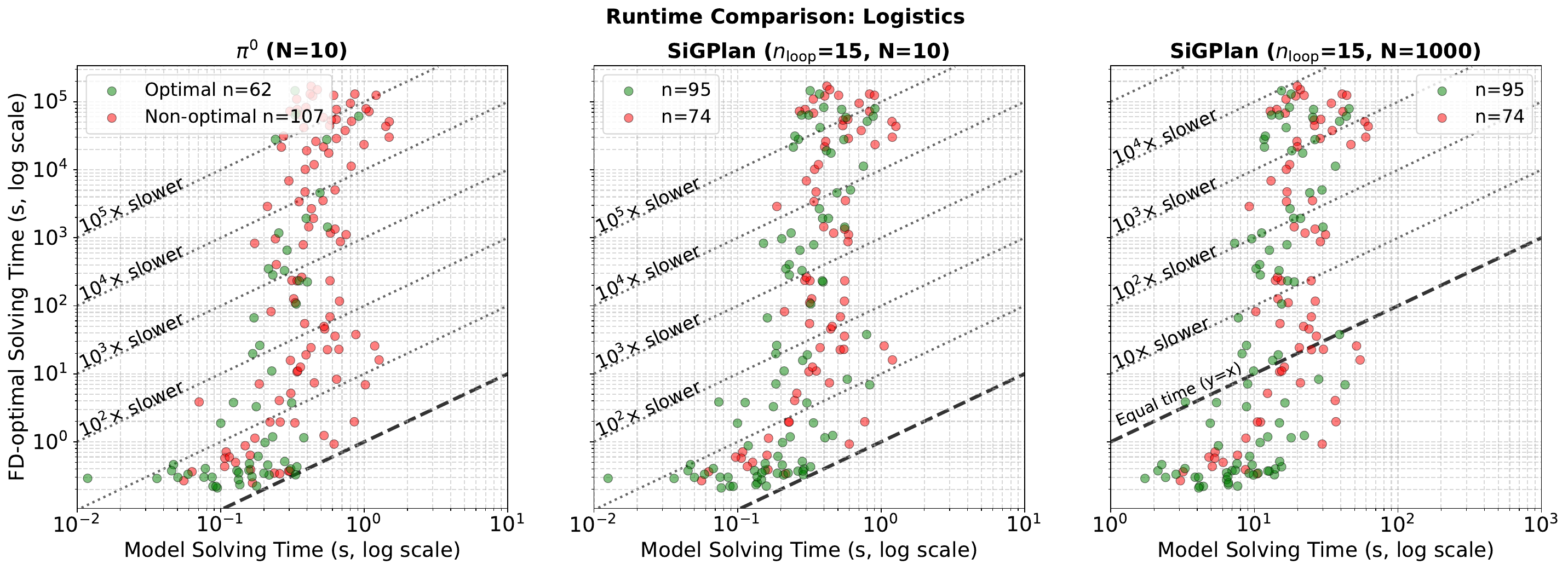}
    \includegraphics[width=0.75\linewidth]{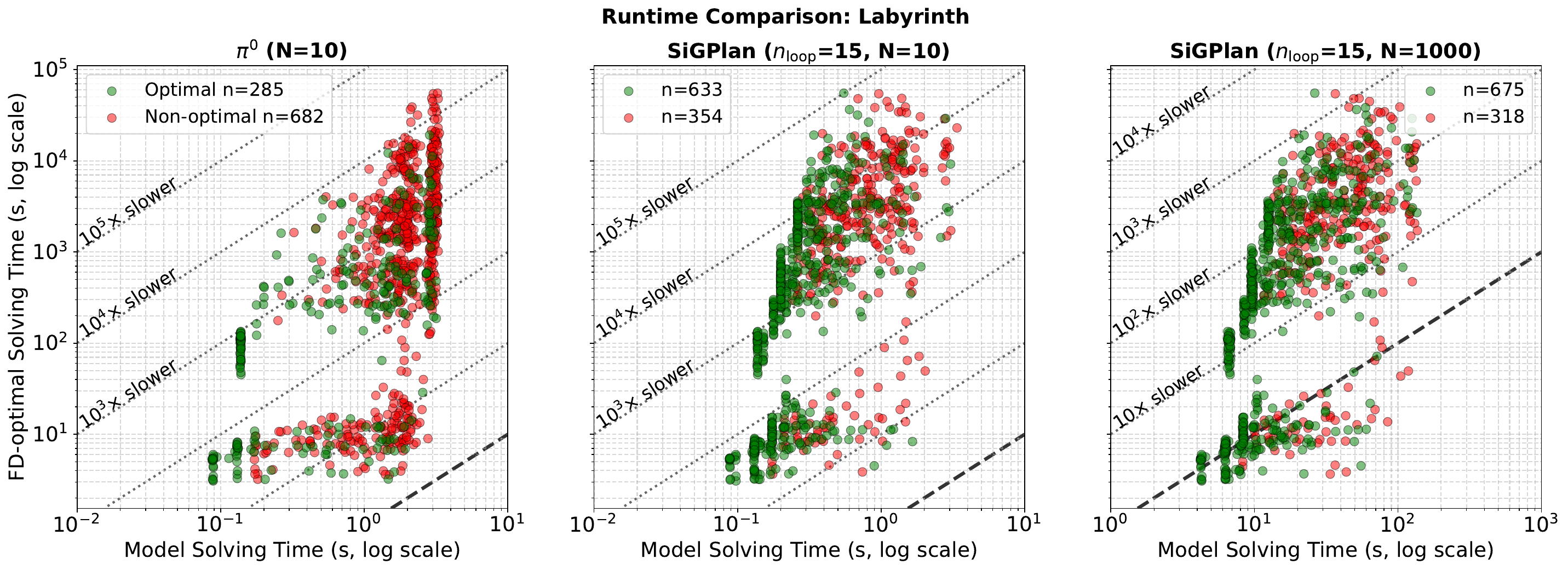}
    \includegraphics[width=0.75\linewidth]{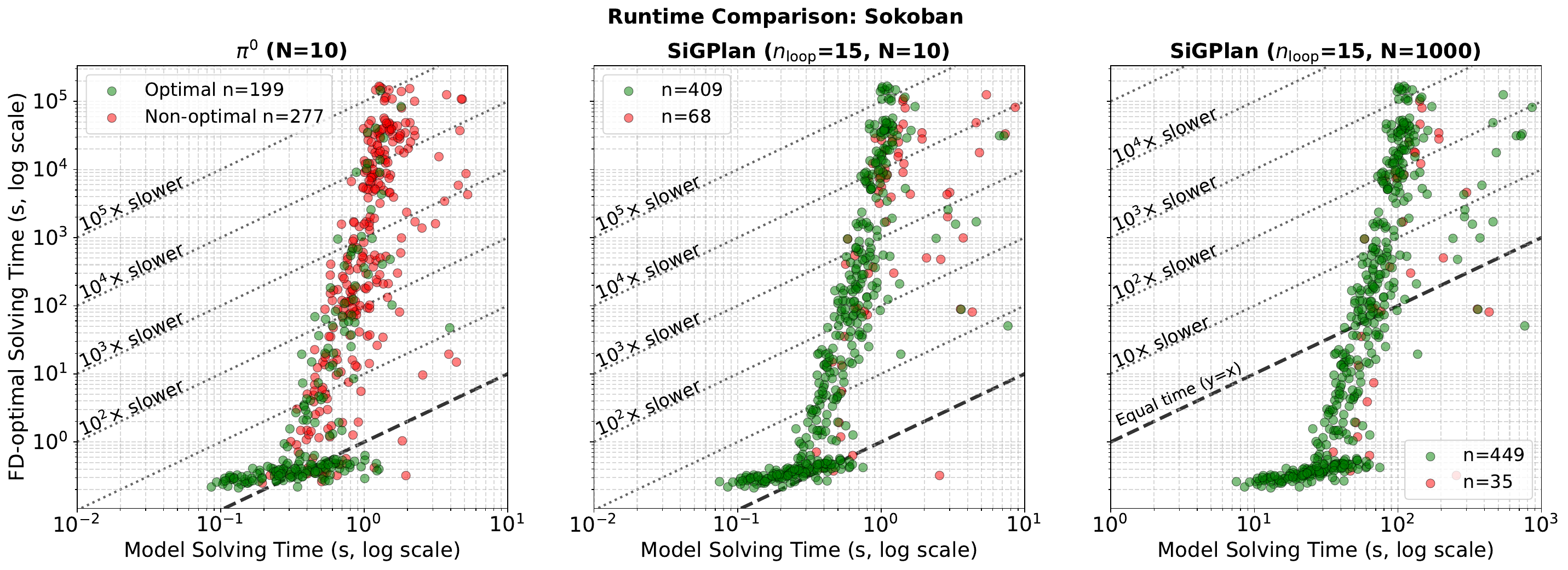}
    \caption{Comparison of solving times (log scale) on all four domains. For the Blocksworld comparison, we ran \emph{FD-optimal} for 5 days on the 1000 test problems, resulting in 643 solved instances.
    For all other domains, \emph{FD-optimal} ran for 48 hours, which led to 169 (Logistics), 1000 (Labyrinth), and 489 (Sokoban)  solved instances, respectively.}
    \label{fig:runtimes_all_domains}
\end{figure*}

Figure~\ref{fig:runtimes_all_domains} extends Figure~\ref{fig:runtime_blockworld} of the main paper. The scatterplots show per-instance comparisons of the runtimes of \emph{FD-optimal} (y-axis) against different ablations of our method (x-axis). Both axes are on a log scale.

\begin{table*}[h]
\centering
\centering \renewcommand{\arraystretch}{1.2}
\resizebox{\textwidth}{!}{%
\begin{tabular}{l|cccc}
\hline
\textbf{Model} & \textbf{Blocksworld} & \textbf{Logistics} & \textbf{Labyrinth} & \textbf{Sokoban}\\
\hline
$\pi^0$ ($N{=}10$) & 0.89 (± 0.02) & 2.45 (± 0.06)& 1.733 (± 0.03) & 3.01 (± 0.10)\\
\emph{SiGPlan} ($n_\text{loop}{=}15$, $N{=}10$) & 0.49 (± 0.01) & 2.15 (± 0.05) & 0.54 (± 0.02) & 4.08 (± 0.20)\\
\emph{SiGPlan} ($n_\text{loop}{=}15$, $N{=}1000$) & 21.21 (± 0.61) & 112.91 (± 2.92) & 27.81 (± 0.85) & 407.66 (± 19.71)\\
\emph{FD-LAMA-first} ($<$20min) & 0.39 (± 0.00)& 7.32 (± 0.79) & 270.15 (± 8.91) & 12.70 (± 2.31)\\
\emph{FD-LAMA-anytime} ($<$20min) & 985.82 (± 14.25) & 765.71 (± 9.25) & 935.28 (± 14.86) & 796.18 (± 17.29) \\
\emph{FD-optimal ($<$48h)} & 13311.40 (± 1648.04) & 16670.74 (± 2702.41) & 3083.48 (± 197.18) & 9762.96 (± 1179.95)\\
\hline
\end{tabular}}
\caption{Solving times in seconds (mean ± std. error) on the  test set for each domain (based on the subset of solved problems for each method).}
\label{tab:solving_times}
\end{table*}

\clearpage
\section{Runtime of the Self-Improvement Loop}

To obtain the finetuned models with \emph{SiGPlan}, we performed 15 iterations of the self-improvement loop for each domain. Table \ref{tab:loop_compute} reports the wall clock time for each loop component, accumulated over all 15 iterations. The \emph{Inference} column shows the total time spent on model calls to generate candidate plans, executed on 100 GPUs in parallel with 48 GB VRAM each. The number of plans that can be generated in parallel depends on the available memory; in our setup, this is 40 plans for Blocksworld, 25 for Logistics and Labyrinth, and 10 for Sokoban. This step scales straightforwardly with additional memory. The \emph{Graph construction + BFS} column reports the time required to process candidate plans into graphs and extract the shortest plans on the CPU, performed on a single machine with 192 CPU cores. The \emph{Training} column reports the accumulated training time over 15 iterations, carried out on 4 machines with 8 NVIDIA A100 GPUs (40 GB VRAM each). Hyperparameters used during training are provided in Section \ref{app:training}. The \emph{Duration of 15 Iterations} column reports the total wall clock time to complete 15 iterations on our hardware. Finally, \emph{GPU hours} reports the total GPU hours accumulated across inference and training (scaled by the number of GPUs).

Blocksworld and Labyrinth completed 15 iterations within a day, Logistics within 2 days, and Sokoban in 3.5 days.

\begin{table*}[h]
\centering \renewcommand{\arraystretch}{1.1}
\resizebox{\textwidth}{!}{%
\begin{tabular}{l|ccccc}
\hline
\textbf{Domain} & \textbf{Inference (h)} & \textbf{Graph construction + BFS (h)} & \textbf{Training (h)} & \textbf{Duration of 15 Iterations (h)} & \textbf{GPU hours (h)} \\
\hline
Blocksworld & 8.13 & 1.75 & 8.42 & 18.3 & 1083\\
Logistics & 16.78 & 5.22 & 16.98 & 38.98 & 2222\\
Labyrinth & 7.22 & 1.75 & 7.98 & 16.95 & 977\\
Sokoban & 19.22 & 7.83 & 56.58 & 83.63 & 3733\\
\hline
\end{tabular}}
\caption{Time spent in the different components of the iterative self-improvement loop  over 15 iterations.}
\label{tab:loop_compute}

\end{table*}

\section{Gap to Optimal Plans}
\begin{figure}[h]
    \centering
    \includegraphics[width=0.7\columnwidth]{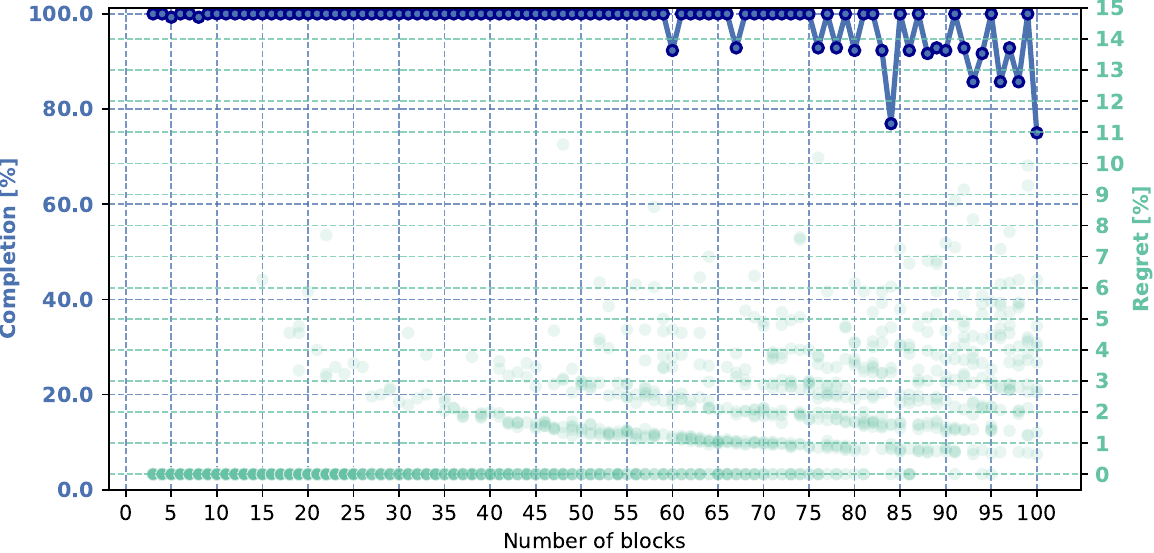}
    \caption{Completion rate (\%, in blue) and regret ($\frac{\text{cost} - \text{cost}_\text{optimal}}{\text{cost}_\text{optimal}} \cdot 100\%$, in green) by block count. Results are shown for the union of the 3 test splits, showing the full block range from 3-100 blocks.}
    \label{fig:regret_scatter}
\end{figure}

Figure \ref{fig:regret_scatter} illustrates the relationships between problem size, completion rate, and regret of a generative model trained on optimal Blocksworld data with up to 100 blocks. It extends Section~\ref{sec:GM:optimality_experiments} of the main paper. The generative model maintains near-perfect completion rates for problems up to approximately 60 blocks, after which occasional dips occur. Regarding solution quality, the regret remains predominantly at 0\% for problems with fewer than 30 blocks, then gradually increases with problem size.

\begin{figure*}[htb]
    \centering
    \includegraphics[width=0.9\linewidth]{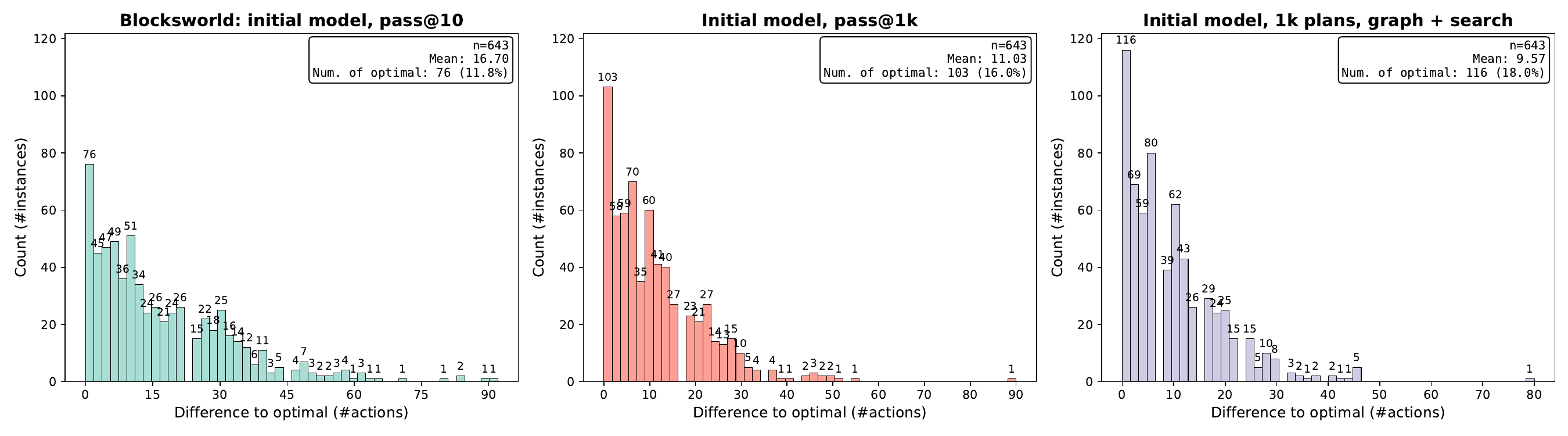}
    \includegraphics[width=0.9\linewidth]{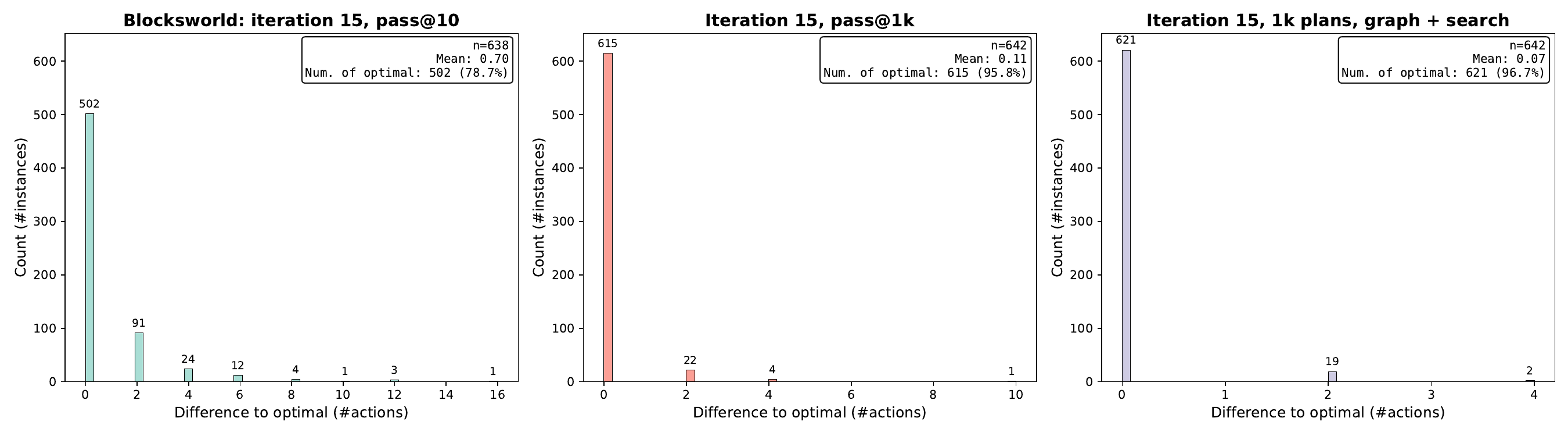}
    \caption{Blocksworld -- Plan length differences between generated plans and optimal solutions. The top row shows performance of the pretrained model, while the bottom row shows the model's performance after 15 iterations of self-improvement. For both, we present findings for different inference parameters.}
    \label{fig:difference_to_optimal_bw}
\end{figure*}

\begin{figure*}[htb]
    \centering
    \includegraphics[width=0.9\linewidth]{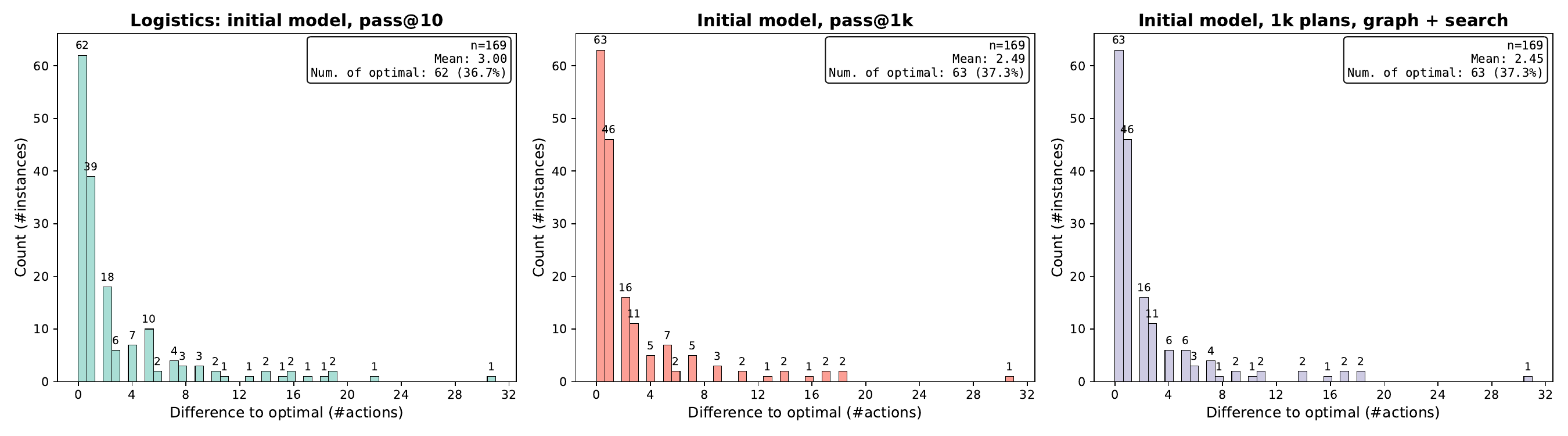}
    \includegraphics[width=0.9\linewidth]{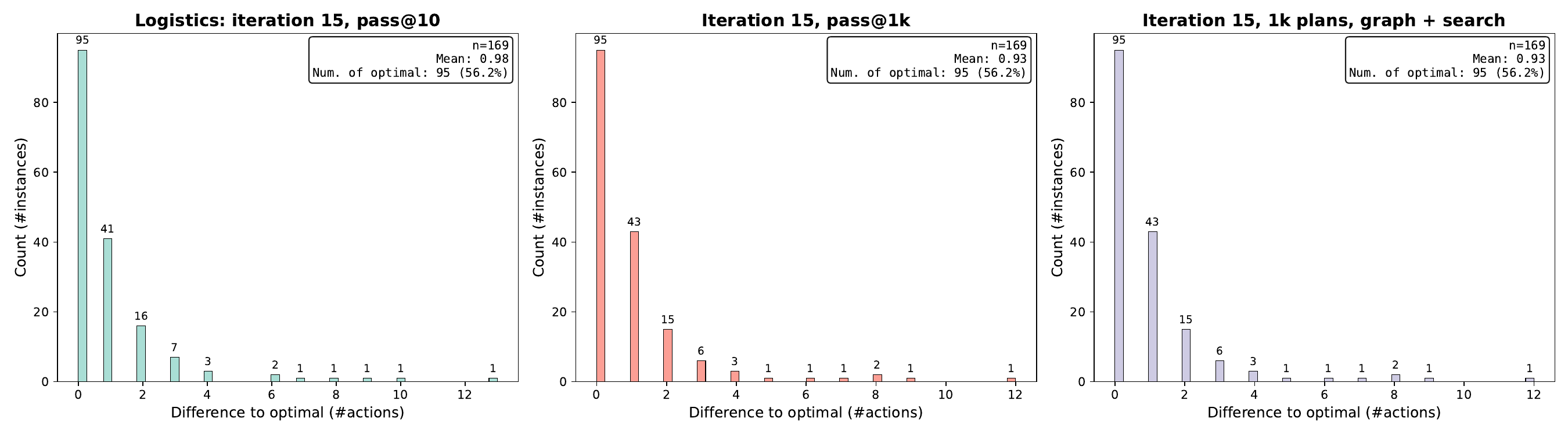}
    \caption{Logistics -- Plan length differences between generated plans and optimal solutions. The top row shows performance of the pretrained model, while the bottom row shows the model's performance after 15 iterations of self-improvement. For both, we present findings for different inference parameters.}
    \label{fig:difference_to_optimal_log}
\end{figure*}

\begin{figure*}[htb]
    \centering
    \includegraphics[width=0.9\linewidth]{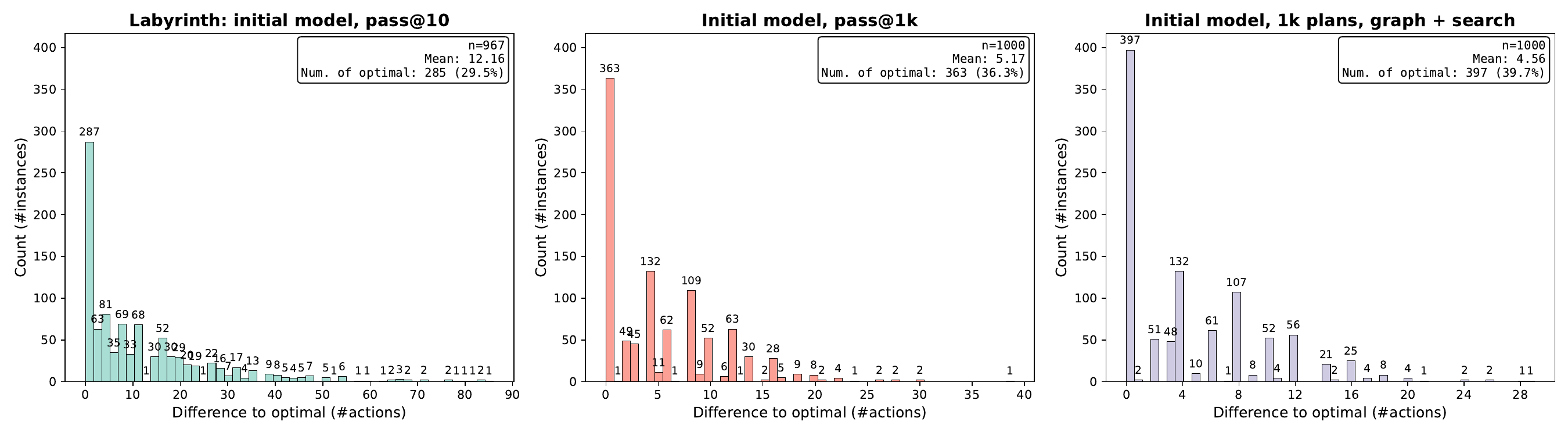}
    \includegraphics[width=0.9\linewidth]{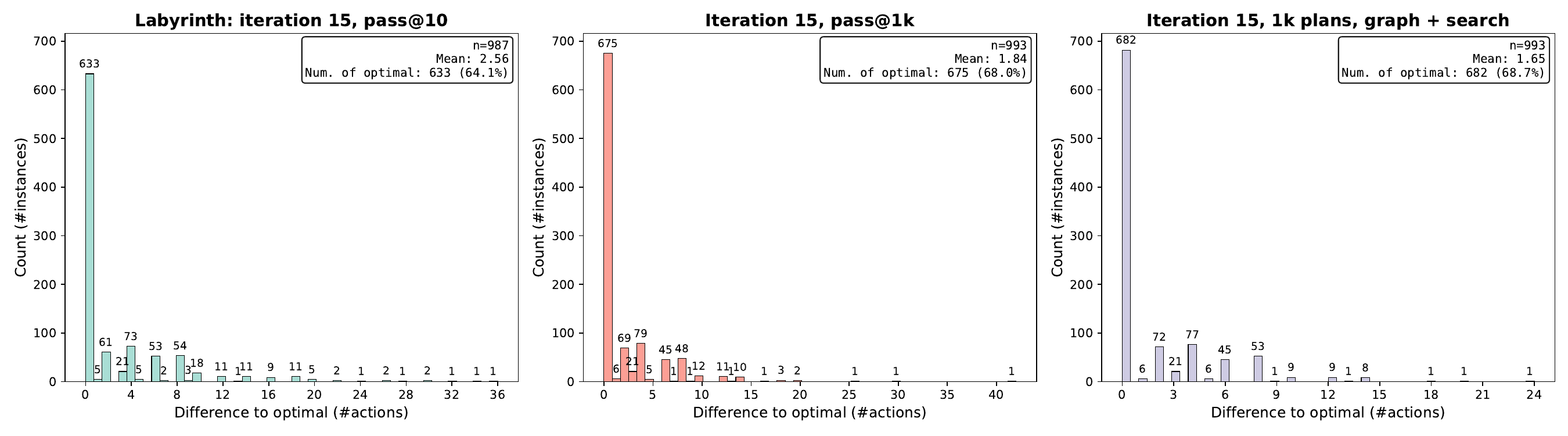}
    \caption{Labyrinth -- Plan length differences between generated plans and optimal solutions. The top row shows performance of the pretrained model, while the bottom row shows the model's performance after 15 iterations of self-improvement. For both, we present findings for different inference parameters.}
    \label{fig:difference_to_optimal_lab}
\end{figure*}

\begin{figure*}[htb]
    \centering
    \includegraphics[width=0.9\linewidth]{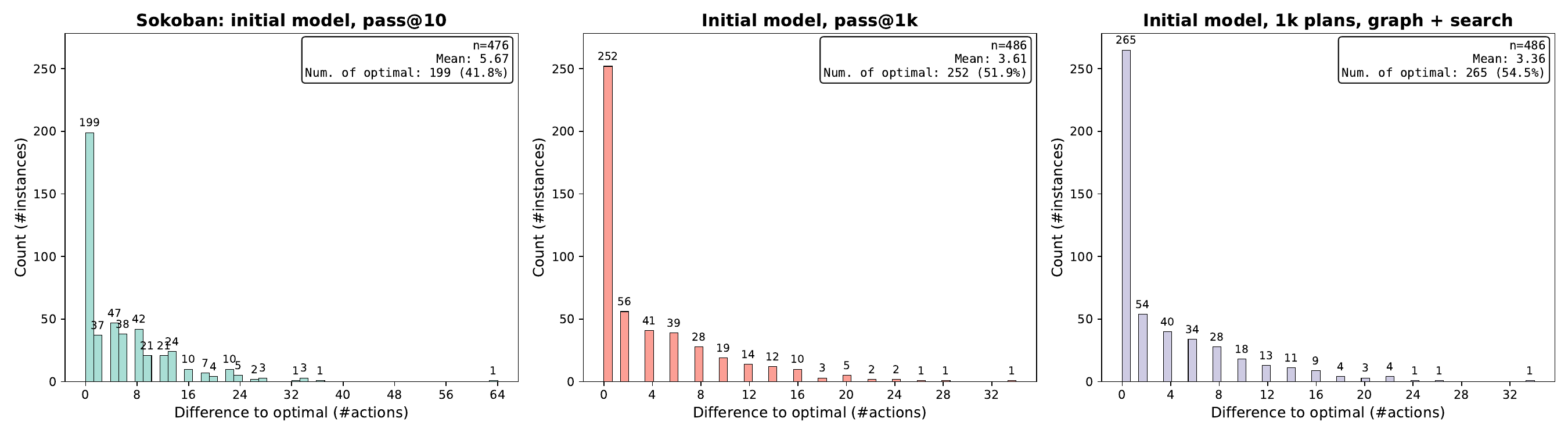}
    \includegraphics[width=0.9\linewidth]{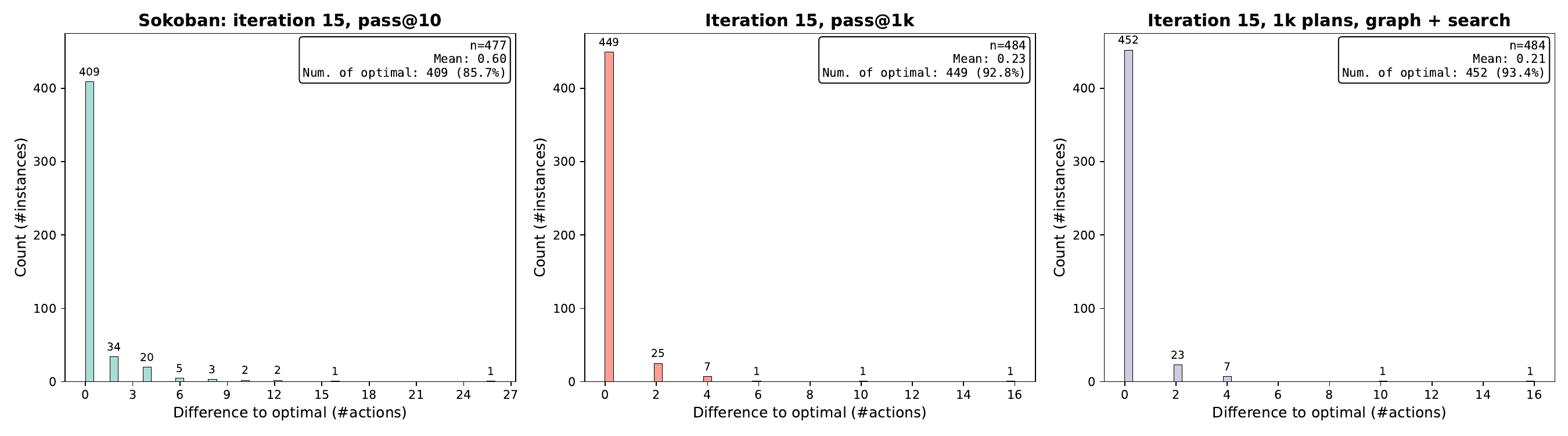}
    \caption{Sokoban -- Plan length differences between generated plans and optimal solutions. The top row shows performance of the pretrained model, while the bottom row shows the model's performance after 15 iterations of self-improvement. For both, we present findings for different inference parameters.}
    \label{fig:difference_to_optimal_sok}
\end{figure*}

In Section~\ref{sec:experiments} of the main paper, we show that \textit{SiGPlan} produces optimal plans in most test cases. As shown there, our Sokoban model generates optimal plans in over 90\% of test cases (where the optimal plan length is known) after the self-improvement loop with $N{=}1000$. Figures~\ref{fig:difference_to_optimal_bw}--\ref{fig:difference_to_optimal_sok} show the corresponding differences to optimal plan length for all four domains. We also include an additional setting with further test-time scaling: in the last column of each figure, we construct a graph from the 1000 candidate plans and perform a BFS to extract the shortest plan present in the graph. This is the same procedure used to generate improved training plans during the self-improvement loop.

\clearpage
\section{Plan Length Distributions}

\begin{figure*}[htb]
    \centering
    \includegraphics[width=0.45\linewidth]{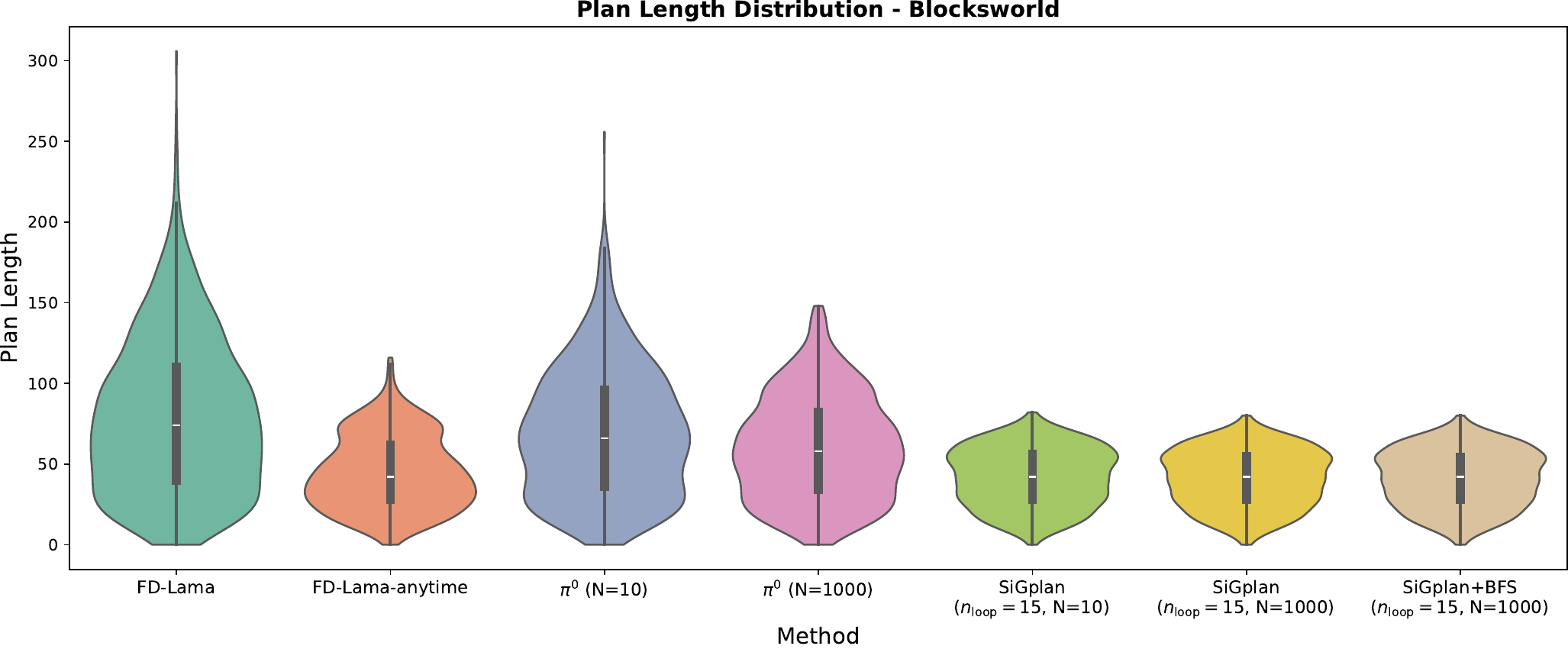}%
    \includegraphics[width=0.45\linewidth]{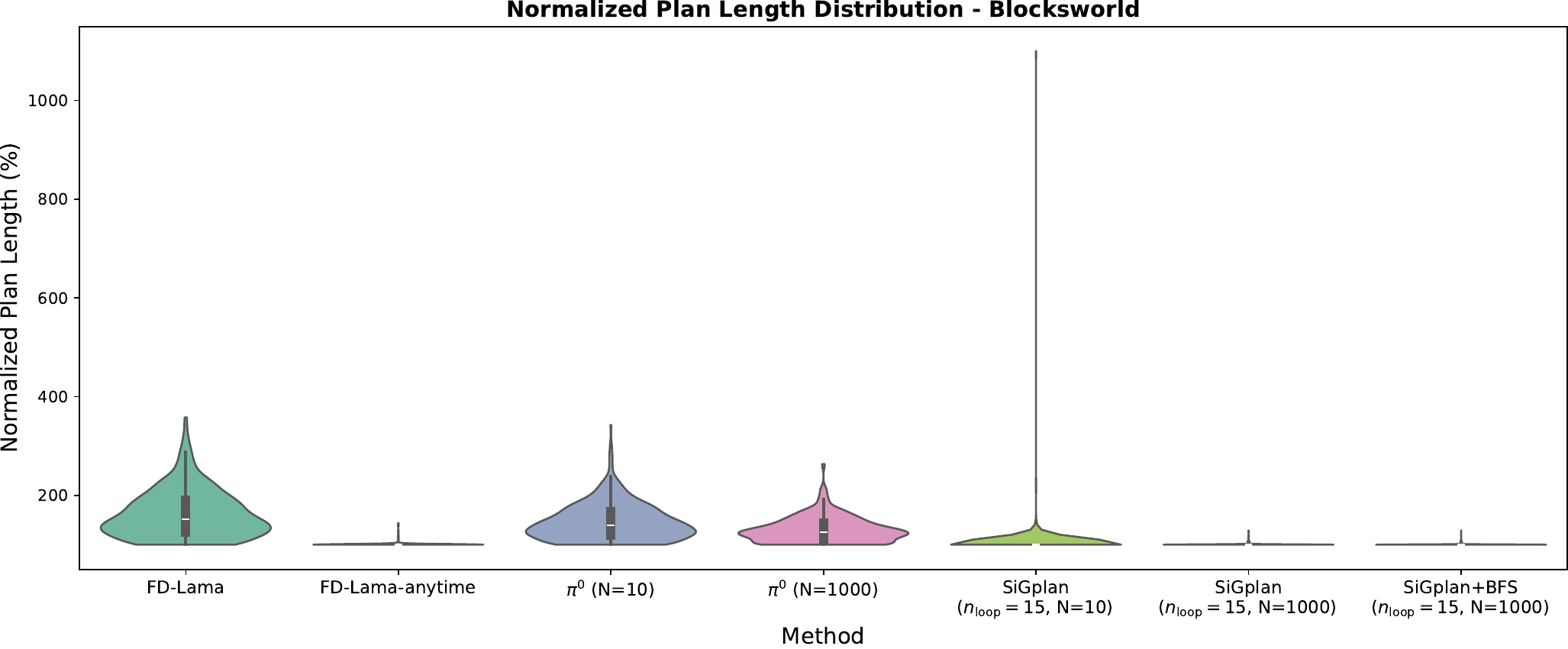}
    
    \vspace{0.3cm}
    
    \includegraphics[width=0.45\linewidth]
    {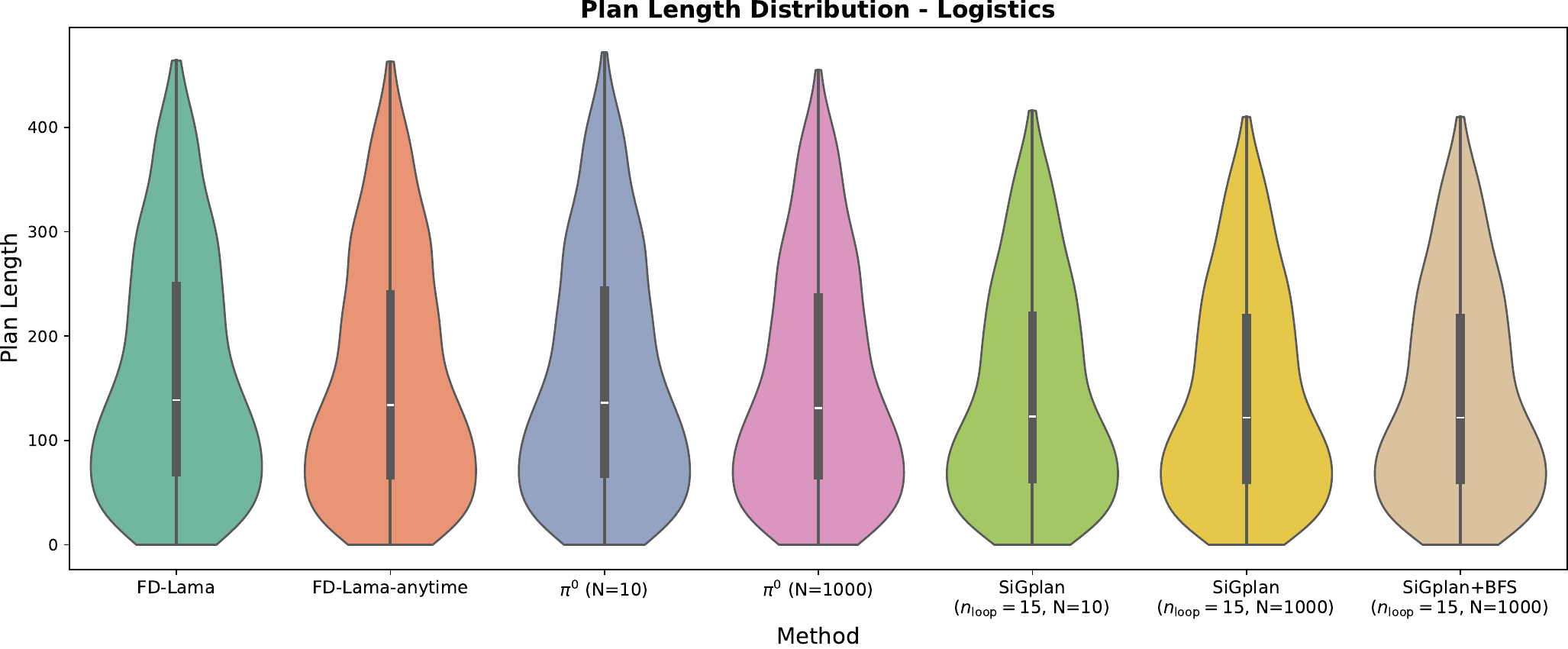}%
    \includegraphics[width=0.45\linewidth]{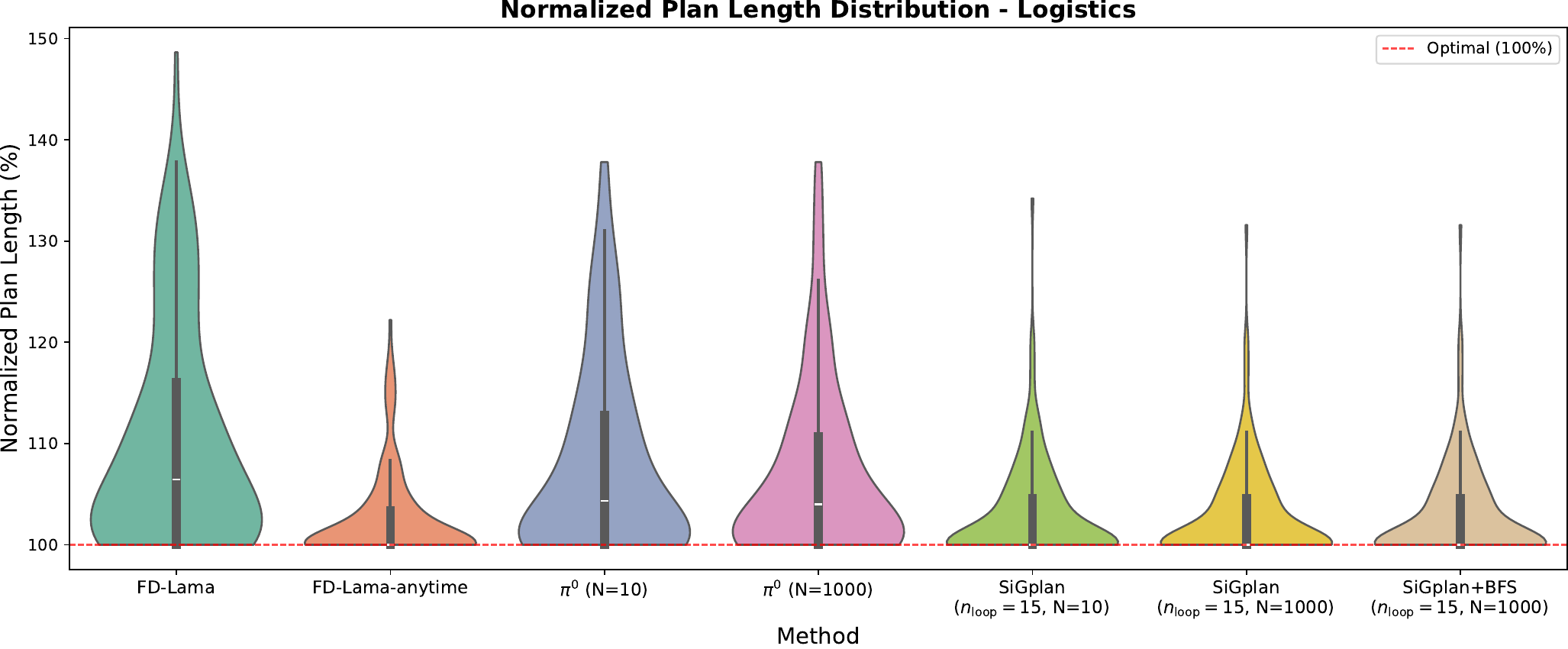}

    \vspace{0.3cm}
    
    \includegraphics[width=0.45\linewidth]{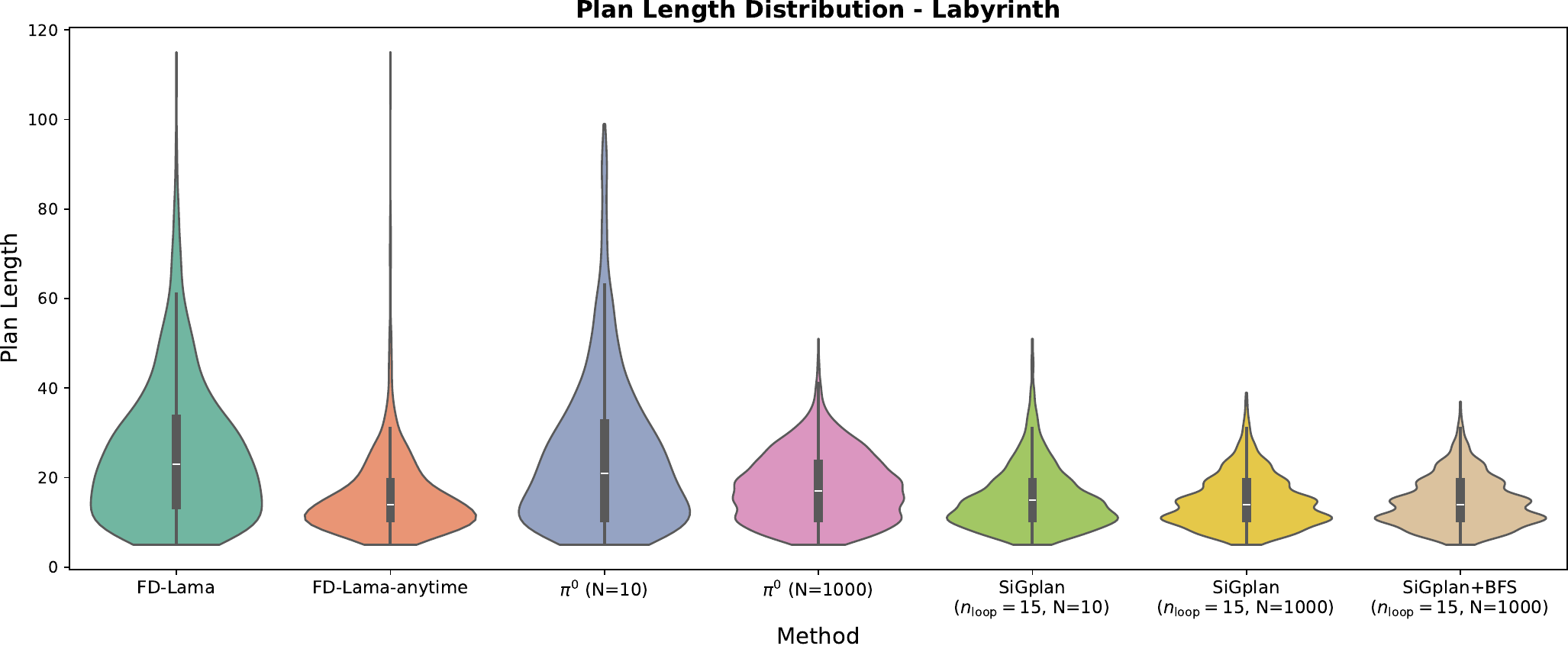}%
    \includegraphics[width=0.45\linewidth]{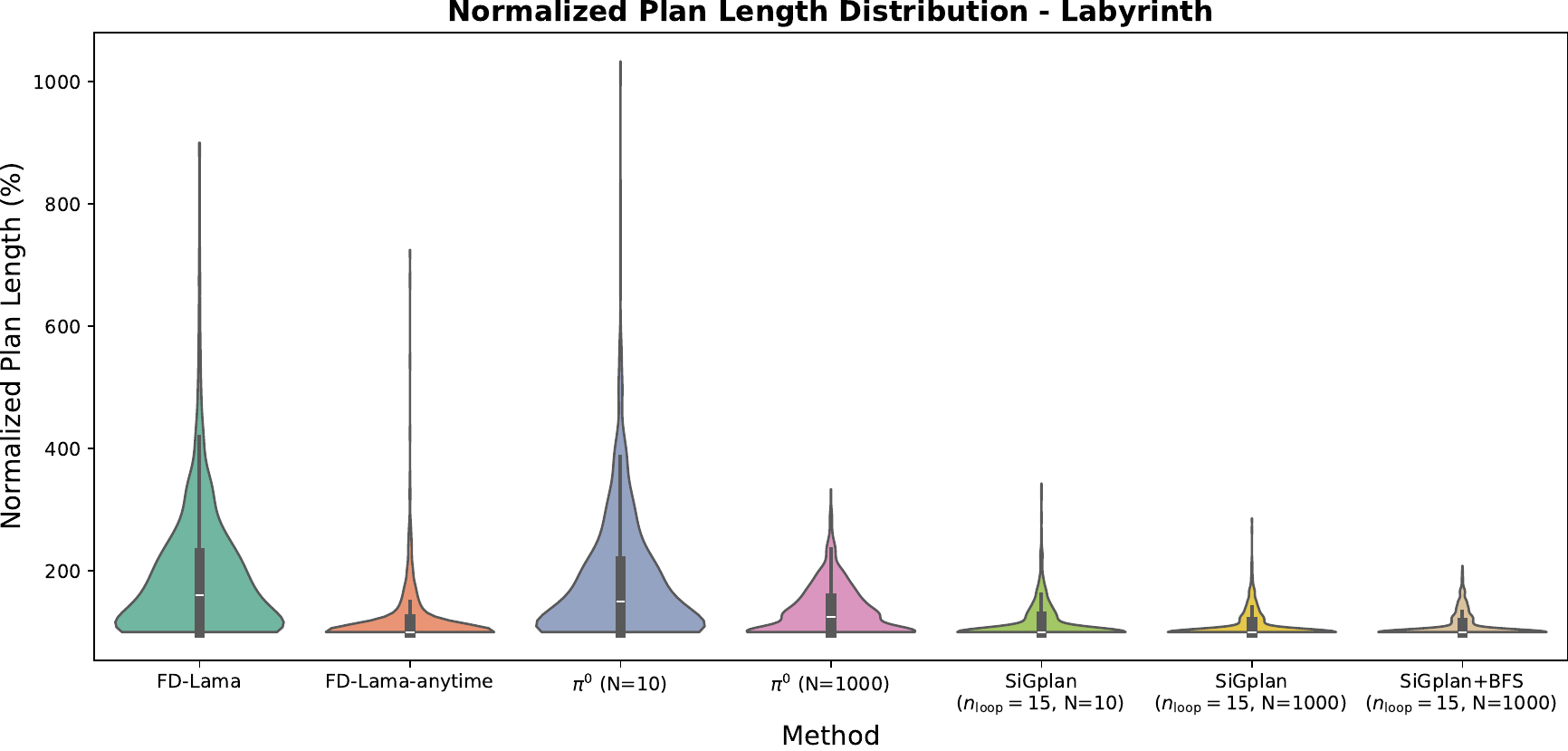}
    
    \vspace{0.3cm}
    
    \includegraphics[width=0.45\linewidth]{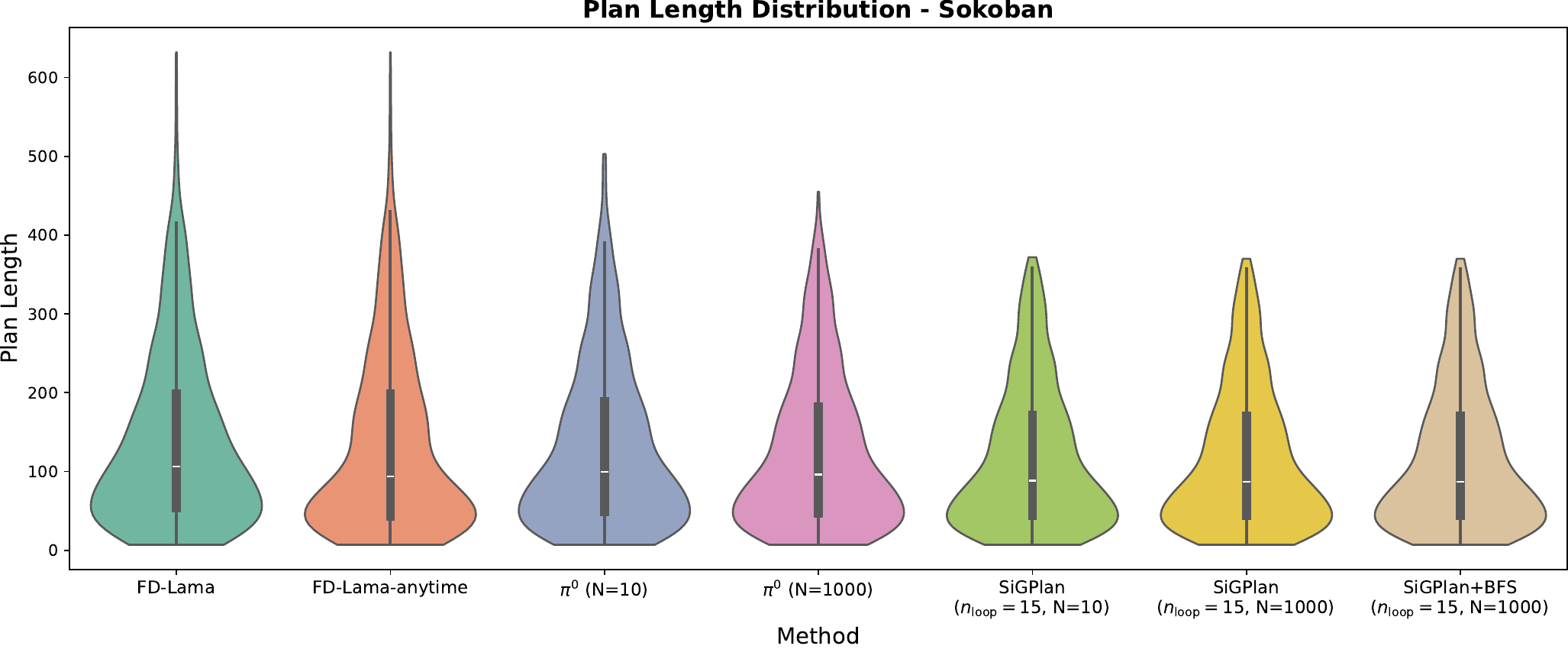}%
    \includegraphics[width=0.45\linewidth]{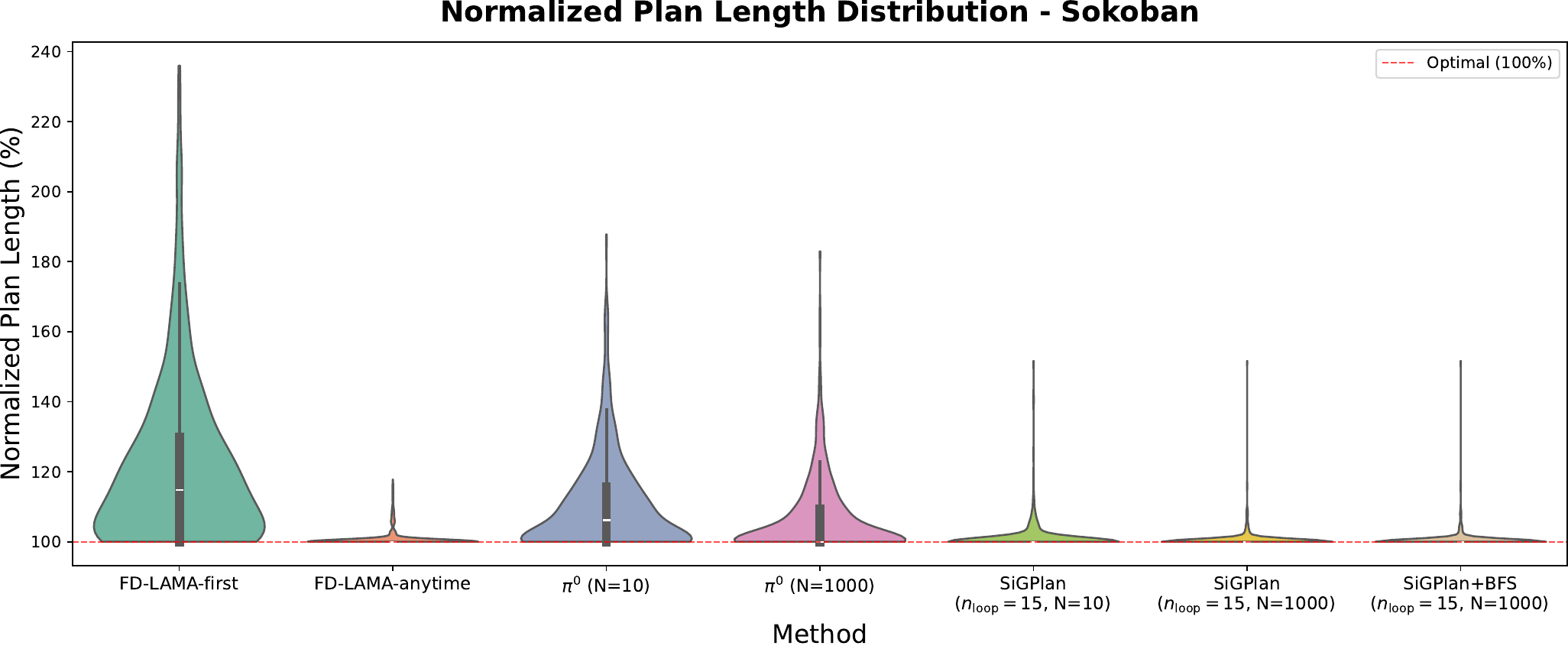}
    
    \caption{Distribution of plan length (left column) and normalized plan length (right column) for different methods across different domains. Normalized plan length expresses the plan length as a percentage of the optimal plan length (100\% = optimal): $\frac{\text{cost} + 1}{\text{cost}_\text{optimal} + 1} \cdot 100\%$, where adding 1 avoids division by zero for trivial problems with optimal plan length 0. The outlier in the normalized plan length plot for the Blocksworld domain corresponds to a trivial problem where the goal conditions are already satisfied in the initial state, and \textit{SiGPlan} ($n_\text{loop}=15$, $N=10$) generated a plan with 10 actions.} 
    \label{fig:violin}
\end{figure*}

\clearpage
\section{Statistical Analysis}

We compared the following conditions: (1) $\pi^0$ vs.~\emph{SiGPlan} with $N{=}10$, (2) $\pi^0$ vs.~\emph{SiGPlan} with $N{=}10^3$, (3) $\pi^0$ with $N{=}10$ vs.~$N{=}10^3$, (4) \emph{SiGPlan} with $N{=}10$ vs.~$N{=}10^3$, (5) \emph{SiGPlan} ($N{=}10^3$) vs.~\emph{SiGPlan+BFS} ($N{=}10^3$), and (6) \emph{SiGPlan+BFS} ($N{=}10^3$) vs.~\emph{FD-LAMA-anytime}.

Prior to analysis, we conducted a Shapiro-Wilk test on our Blocksworld results, which revealed that the paired differences in plan length significantly deviate from normality across all comparisons (all $p < 0.001$).

For the plan length analysis, we therefore chose the non-parametric Wilcoxon signed-rank test rather than a paired t-test, applied consistently across all domains. The Wilcoxon signed-rank test is a conservative method requiring fewer assumptions about the data distribution than a parametric t-test, while maintaining statistical power.

Since plan length can only be compared when all methods successfully generate a plan, we restricted the analysis to problems where all methods under comparison found valid plans. This resulted in sample sizes of 992 (Blocksworld), 986 (Logistics), 955 (Labyrinth), and 910 (Sokoban) problems, respectively. All p-values reported in the main paper are Bonferroni-corrected for six comparisons (adjusted $\alpha$-level: 0.008).

For completion rate, we applied McNemar's test to all 1000 problems per domain, as this metric is defined for all problems regardless of solution status. Again, all reported p-values are Bonferroni-corrected for six comparisons (adjusted $\alpha$-level: 0.008).
\clearpage
\section{Domains} \label{app:domains}

For Blocksworld, Logistics, and Sokoban, domain files and problem instance generators were generated using the PDDL generator collection from \citet{seipp2022pddlgen}.

\subsection{Blocksworld}

\begin{lstlisting}[caption=Blocksworld domain file]
(define (domain blocksworld-4ops)
  (:requirements :strips)
(:predicates (clear ?x)
             (on-table ?x)
             (arm-empty)
             (holding ?x)
             (on ?x ?y))

(:action pickup
  :parameters (?ob)
  :precondition (and (clear ?ob) (on-table ?ob) (arm-empty))
  :effect (and (holding ?ob) (not (clear ?ob)) (not (on-table ?ob)) 
               (not (arm-empty))))

(:action putdown
  :parameters  (?ob)
  :precondition (holding ?ob)
  :effect (and (clear ?ob) (arm-empty) (on-table ?ob) 
               (not (holding ?ob))))

(:action stack
  :parameters  (?ob ?underob)
  :precondition (and (clear ?underob) (holding ?ob))
  :effect (and (arm-empty) (clear ?ob) (on ?ob ?underob)
               (not (clear ?underob)) (not (holding ?ob))))

(:action unstack
  :parameters  (?ob ?underob)
  :precondition (and (on ?ob ?underob) (clear ?ob) (arm-empty))
  :effect (and (holding ?ob) (clear ?underob)
               (not (on ?ob ?underob)) (not (clear ?ob)) (not (arm-empty)))))
\end{lstlisting}

\clearpage
\subsection{Logistics}

\begin{lstlisting}[caption=Logistics domain file]
(define (domain logistics)
  (:requirements :strips :typing)
  (:types 
  	package location vehicle - object
  	truck airplane - vehicle
  	city airport - location)
  
  (:predicates 	
		(at ?vehicle-or-package - (either vehicle package)  ?location - location)
		(in ?package - package ?vehicle - vehicle)
		(in-city ?loc-or-truck - (either location truck) ?citys - city))
		
  (:action load-truck
	:parameters
		 (?obj - package
		  ?truck - truck
		  ?loc - location)
	:precondition
		(and 	(at ?truck ?loc) 
			(at ?obj ?loc))
	:effect
		(and 	(not (at ?obj ?loc)) 
			(in ?obj ?truck)))

  (:action load-airplane
	:parameters
		(?obj - package
		 ?airplane - airplane
		 ?loc - airport)
	:precondition
		(and
			(at ?obj ?loc) 
			(at ?airplane ?loc))
	:effect
   		(and 	(not (at ?obj ?loc)) 
			(in ?obj ?airplane)))

  (:action unload-truck
	:parameters
		(?obj - package
		 ?truck - truck
		 ?loc - location)
	:precondition
		(and    (at ?truck ?loc) 
			(in ?obj ?truck))
	:effect
		(and	(not (in ?obj ?truck)) 
			(at ?obj ?loc)))

  (:action unload-airplane
	:parameters
		(?obj - package
		 ?airplane - airplane
		 ?loc - airport)
	:precondition
		(and	(in ?obj ?airplane) 
			(at ?airplane ?loc))
	:effect
		(and 
			(not (in ?obj ?airplane)) 
			(at ?obj ?loc)))

  (:action drive-truck
	:parameters
		(?truck - truck
		 ?loc-from - location
		 ?loc-to - location
		 ?city - city)
	:precondition
		(and 	(at ?truck ?loc-from)
			(in-city ?loc-from ?city)
			(in-city ?loc-to ?city))
	:effect
		(and 	(not (at ?truck ?loc-from)) 
			(at ?truck ?loc-to)))

  (:action fly-airplane
	:parameters
		(?airplane - airplane
		 ?loc-from - airport
		 ?loc-to - airport)
	:precondition
		(at ?airplane ?loc-from)
	:effect
		(and 	(not (at ?airplane ?loc-from)) 
		(at ?airplane ?loc-to)))
)
\end{lstlisting}

\clearpage
\subsection{Labyrinth}
The Labyrinth domain was part of the Satisficing track of the most recent International Planning Competition (IPC) 2023 \cite{eifler2023labyrinth}. 
We chose this domain as it was shown to be particularly challenging for the competing planners at the IPC 2023 \cite{taitler2023ipc}. The original domain defined a cost of 0 for move-card-$\langle$direction$\rangle$ and end-move-card-$\langle$direction$\rangle$ actions. Since we define plan length by the number of actions in a plan, we adapted the domain definition to add a cost of 1 to every action for optimal plan generation with Fast Downward.

\begin{lstlisting}[caption=Labyrinth domain file]
(define (domain labyrinth)
(:requirements :adl :action-costs :strips :typing :numeric-fluents)

(:types
    card - object
    direction - object
    directionV - direction
    directionH - direction
    gridpos - object
)

(:constants
    s n - directionV
    w e - directionH
)

(:predicates
    (next ?p1 - gridpos ?p2 - gridpos)
    (max-pos ?p - gridpos)
    (min-pos ?p - gridpos)
    (blocked ?c - card ?d - direction)
    (robot-at ?c - card)
    (card-at ?c - card ?x - gridpos ?y - gridpos)
    (left)
    (cards-moving)
    (cards-moving-west)
    (cards-moving-east)
    (cards-moving-south)
    (cards-moving-north)
    (next-moving-card ?c - card)
    (new-headtail-card ?c - card)

)

(:functions
    (total-cost) - number
    (move-robot-cost) - number
    (move-card) - number
)

(:action move-west
    :parameters (?cfrom - card ?xfrom - gridpos ?yfrom - gridpos ?dfrom - directionH ?cto - card ?xto - gridpos ?yto - gridpos ?dto - directionH)
    :precondition
        (and
            (not (cards-moving))
            (= ?dfrom w)
            (robot-at ?cfrom)
            (card-at ?cfrom ?xfrom ?yfrom)
            (card-at ?cto ?xto ?yto)
            (next ?xfrom ?xto)
            (= ?yfrom ?yto)
            (not (= ?dfrom ?dto))
            (not (blocked ?cfrom ?dfrom))
            (not (blocked ?cto ?dto))
        )
    :effect
        (and
            (not (robot-at ?cfrom))
            (robot-at ?cto)
            (increase (total-cost) (move-robot-cost))
        )
)

(:action move-east
    :parameters (?cfrom - card ?xfrom - gridpos ?yfrom - gridpos ?dfrom - directionH ?cto - card ?xto - gridpos ?yto - gridpos ?dto - directionH)
    :precondition
        (and
            (not (cards-moving))
            (= ?dfrom e)
            (robot-at ?cfrom)
            (card-at ?cfrom ?xfrom ?yfrom)
            (card-at ?cto ?xto ?yto)
            (next ?xto ?xfrom)
            (= ?yfrom ?yto)
            (not (= ?dfrom ?dto))
            (not (blocked ?cfrom ?dfrom))
            (not (blocked ?cto ?dto))
        )
    :effect
        (and
            (not (robot-at ?cfrom))
            (robot-at ?cto)
            (increase (total-cost) (move-robot-cost))
        )
)

(:action move-north
    :parameters (?cfrom - card ?xfrom - gridpos ?yfrom - gridpos ?dfrom - directionV ?cto - card ?xto - gridpos ?yto - gridpos ?dto - directionV)
    :precondition
        (and
            (not (cards-moving))
            (= ?dfrom n)
            (robot-at ?cfrom)
            (card-at ?cfrom ?xfrom ?yfrom)
            (card-at ?cto ?xto ?yto)
            (next ?yfrom ?yto)
            (= ?xfrom ?xto)
            (not (= ?dfrom ?dto))
            (not (blocked ?cfrom ?dfrom))
            (not (blocked ?cto ?dto))
        )
    :effect
        (and
            (not (robot-at ?cfrom))
            (robot-at ?cto)
            (increase (total-cost) (move-robot-cost))
        )
)

(:action move-south
    :parameters (?cfrom - card ?xfrom - gridpos ?yfrom - gridpos ?dfrom - directionV ?cto - card ?xto - gridpos ?yto - gridpos ?dto - directionV)
    :precondition
        (and
            (not (cards-moving))
            (= ?dfrom s)
            (robot-at ?cfrom)
            (card-at ?cfrom ?xfrom ?yfrom)
            (card-at ?cto ?xto ?yto)
            (next ?yto ?yfrom)
            (= ?xfrom ?xto)
            (not (= ?dfrom ?dto))
            (not (blocked ?cfrom ?dfrom))
            (not (blocked ?cto ?dto))
        )
    :effect
        (and
            (not (robot-at ?cfrom))
            (robot-at ?cto)
            (increase (total-cost) (move-robot-cost))
        )
)

(:action start-move-card-west
:parameters(?cm - card ?x - gridpos ?y - gridpos ?cnext - card ?nextx - gridpos)
:precondition
    (and
        (not (cards-moving))
        (not (cards-moving-west))
        (not (robot-at ?cm))
        (card-at ?cm ?x ?y )
        (min-pos ?x)
        (card-at ?cnext ?nextx ?y)
        (next ?nextx ?x)
    )
:effect
    (and
        (cards-moving)
        (cards-moving-west)
        (not (card-at ?cm ?x ?y ))
        (new-headtail-card ?cm)
        (next-moving-card ?cnext)
        (increase (total-cost) (move-card))
    )
)

(:action move-card-west
:parameters(?cm - card ?x - gridpos ?y - gridpos ?cnext - card ?nextx - gridpos ?prevx - gridpos)
:precondition
    (and
        (cards-moving)
        (cards-moving-west)
        (not (robot-at ?cm))
        (next-moving-card ?cm)
        (card-at ?cm ?x ?y )
        (card-at ?cnext ?nextx ?y)
        (next ?x ?prevx)
        (next ?nextx ?x)
    )
:effect
    (and
        (cards-moving)
        (cards-moving-west)
        (not (card-at ?cm ?x ?y))
        (card-at ?cm ?prevx ?y)
        (not (next-moving-card ?cm))
        (next-moving-card ?cnext)
        (increase (total-cost) (move-card))
    )
)

(:action stop-move-card-west
:parameters(?cm - card ?x - gridpos ?y - gridpos ?prevx - gridpos ?newtc - card)
:precondition
    (and
        (cards-moving)
        (cards-moving-west)
        (not (robot-at ?cm))
        (next-moving-card ?cm)
        (card-at ?cm ?x ?y )
        (next ?x ?prevx)
        (max-pos ?x)
        (new-headtail-card ?newtc)
    )
:effect
    (and
        (not (cards-moving))
        (not (cards-moving-west))
        (not (card-at ?cm ?x ?y))
        (card-at ?cm ?prevx ?y)
        (card-at ?newtc ?x ?y)
        (not (new-headtail-card ?newtc))
        (not (next-moving-card ?cm))
        (increase (total-cost) (move-card)) 
    )
)

(:action start-move-card-east
:parameters(?cm - card ?x - gridpos ?y - gridpos ?cnext - card ?nextx - gridpos)
:precondition
    (and
        (not (cards-moving))
        (not (cards-moving-east))
        (not (robot-at ?cm))
        (card-at ?cm ?x ?y )
        (max-pos ?x)
        (card-at ?cnext ?nextx ?y)
        (next ?x ?nextx)
    )
:effect
    (and
        (cards-moving)
        (cards-moving-east)
        (not (card-at ?cm ?x ?y ))
        (new-headtail-card ?cm)
        (next-moving-card ?cnext)
        (increase (total-cost) (move-card))
    )
)

(:action move-card-east
:parameters(?cm - card ?x - gridpos ?y - gridpos ?cnext - card ?nextx - gridpos ?prevx - gridpos)
:precondition
    (and
        (cards-moving)
        (cards-moving-east)
        (not (robot-at ?cm))
        (next-moving-card ?cm)
        (card-at ?cm ?x ?y )
        (card-at ?cnext ?nextx ?y)
        (next ?prevx ?x)
        (next ?x ?nextx)
    )
:effect
    (and
        (cards-moving)
        (cards-moving-east)
        (not (card-at ?cm ?x ?y))
        (card-at ?cm ?prevx ?y)
        (not (next-moving-card ?cm))
        (next-moving-card ?cnext)
        (increase (total-cost) (move-card))
    )
)

(:action stop-move-card-east
:parameters(?cm - card ?x - gridpos ?y - gridpos ?prevx - gridpos ?newtc - card)
:precondition
    (and
        (cards-moving)
        (cards-moving-east)
        (not (robot-at ?cm))
        (next-moving-card ?cm)
        (card-at ?cm ?x ?y )
        (next  ?prevx ?x)
        (min-pos ?x)
        (new-headtail-card ?newtc)
    )
:effect
    (and
        (not (cards-moving))
        (not (cards-moving-east))
        (not (card-at ?cm ?x ?y))
        (card-at ?cm ?prevx ?y)
        (card-at ?newtc ?x ?y)
        (not (new-headtail-card ?newtc))
        (not (next-moving-card ?cm))
        (increase (total-cost) (move-card))
    )
)

(:action start-move-card-north
:parameters(?cm - card ?x - gridpos ?y - gridpos ?cnext - card ?nexty - gridpos)
:precondition
    (and
        (not (cards-moving))
        (not (cards-moving-north))
        (not (robot-at ?cm))
        (card-at ?cm ?x ?y )
        (min-pos ?y)
        (card-at ?cnext ?x ?nexty)
        (next ?nexty ?y)
    )
:effect
    (and
        (cards-moving)
        (cards-moving-north)
        (not (card-at ?cm ?x ?y ))
        (new-headtail-card ?cm)
        (next-moving-card ?cnext)
        (increase (total-cost) (move-card))
    )
)

(:action move-card-north
:parameters(?cm - card ?x - gridpos ?y - gridpos  ?cnext - card ?nexty - gridpos ?prevy - gridpos)
:precondition
    (and
        (cards-moving)
        (cards-moving-north)
        (not (robot-at ?cm))
        (next-moving-card ?cm)
        (card-at ?cm ?x ?y )
        (card-at ?cnext ?x ?nexty)
        (next ?y ?prevy)
        (next ?nexty ?y)
    )
:effect
    (and
        (cards-moving)
        (cards-moving-north)
        (not (card-at ?cm ?x ?y))
        (card-at ?cm ?x ?prevy)
        (not (next-moving-card ?cm))
        (next-moving-card ?cnext)
        (increase (total-cost) (move-card))
    )
)

(:action stop-move-card-north
:parameters(?cm - card ?x - gridpos ?y - gridpos ?prevy - gridpos ?newtc - card)
:precondition
    (and
        (cards-moving)
        (cards-moving-north)
        (not (robot-at ?cm))
        (next-moving-card ?cm)
        (card-at ?cm ?x ?y )
        (next ?y ?prevy)
        (max-pos ?y)
        (new-headtail-card ?newtc)
    )
:effect
    (and
        (not (cards-moving))
        (not (cards-moving-north))
        (not (card-at ?cm ?x ?y))
        (card-at ?cm ?x ?prevy)
        (card-at ?newtc ?x ?y)
        (not (new-headtail-card ?newtc))
        (not (next-moving-card ?cm))
        (increase (total-cost) (move-card))
    )
)

(:action start-move-card-south
:parameters(?cm - card ?x - gridpos ?y - gridpos  ?cnext - card ?nexty - gridpos)
:precondition
    (and
        (not (cards-moving))
        (not (cards-moving-south))
        (not (robot-at ?cm))
        (card-at ?cm ?x ?y )
        (max-pos ?y)
        (card-at ?cnext ?x ?nexty)
        (next ?y ?nexty)
    )
:effect
    (and
        (cards-moving)
        (cards-moving-south)
        (not (card-at ?cm ?x ?y ))
        (new-headtail-card ?cm)
        (next-moving-card ?cnext)
        (increase (total-cost) (move-card))
    )
)

(:action move-card-south
:parameters(?cm - card ?x - gridpos ?y - gridpos  ?cnext - card ?nexty - gridpos ?prevy - gridpos)
:precondition
    (and
        (cards-moving)
        (cards-moving-south)
        (not (robot-at ?cm))
        (next-moving-card ?cm)
        (card-at ?cm ?x ?y )
        (card-at ?cnext ?x ?nexty)
        (next ?prevy ?y)
        (next ?y ?nexty)
    )
:effect
    (and
        (cards-moving)
        (cards-moving-south)
        (not (card-at ?cm ?x ?y))
        (card-at ?cm ?x ?prevy)
        (not (next-moving-card ?cm))
        (next-moving-card ?cnext)
        (increase (total-cost) (move-card))
    )
)

(:action stop-move-card-south
:parameters(?cm - card ?x - gridpos ?y - gridpos ?prevy - gridpos ?newtc - card)
:precondition
    (and
        (cards-moving)
        (cards-moving-south)
        (not (robot-at ?cm))
        (next-moving-card ?cm)
        (card-at ?cm ?x ?y )
        (next ?prevy ?y)
        (min-pos ?y)
        (new-headtail-card ?newtc)
    )
:effect
    (and
        (not (cards-moving))
        (not (cards-moving-south))
        (not (card-at ?cm ?x ?y))
        (card-at ?cm ?x ?prevy)
        (card-at ?newtc ?x ?y)
        (not (new-headtail-card ?newtc))
        (not (next-moving-card ?cm))
        (increase (total-cost) (move-card))
    )
)

(:action leave
:parameters(?c - card ?prow - gridpos ?pcolumn - gridpos)
:precondition
    (and
        (not (cards-moving))
        (robot-at ?c)
        (card-at ?c ?prow ?pcolumn)
        (max-pos ?prow)
        (max-pos ?pcolumn)
        (not (blocked ?c s ))
    )
:effect
    (and
        (increase (total-cost) (move-card))
        (left)
    )
)
)

\end{lstlisting}

\subsection{Sokoban}
We slightly modified the Sokoban domain \cite{seipp2022pddlgen} by adding an additional predicate \texttt{(has-box ?l - LOC)}. With this predicate, we specify goal conditions through a set of target positions for boxes, but without an explicit assignment of boxes to those positions. This definition is in line with other state-of-the-art Sokoban solvers like the Sokolution and Festival solvers.

\begin{lstlisting}[caption=Sokoban domain file]
(define (domain typed-sokoban)
(:requirements :typing)
(:types LOC DIR)
(:predicates 
             (at-robot ?l - LOC)
             (has-box ?l - LOC)
             (adjacent ?l1 - LOC ?l2 - LOC ?d - DIR) 
             (clear ?l - LOC)
)


(:action move
:parameters (?from - LOC ?to - LOC ?dir - DIR)
:precondition (and (clear ?to) (at-robot ?from) (adjacent ?from ?to ?dir))
:effect (and (at-robot ?to) (not (at-robot ?from)))
)
             

(:action push
:parameters  (?rloc - LOC ?bloc - LOC ?floc - LOC ?dir - DIR)
:precondition (and (at-robot ?rloc) (has-box ?bloc) (clear ?floc)
	           (adjacent ?rloc ?bloc ?dir) (adjacent ?bloc ?floc ?dir))

:effect (and (at-robot ?bloc) (has-box ?floc) (clear ?bloc)
             (not (at-robot ?rloc)) (not (has-box ?bloc)) (not (clear ?floc)))
)
\end{lstlisting}

\end{document}